\documentclass[lettersize,journal]{IEEEtran}
\usepackage{amsmath,amsfonts}
\usepackage{amsthm}
\usepackage{amssymb}
\newtheorem{thm}{Theorem}

\usepackage{algorithmic}
\usepackage{algorithm}
\usepackage{array}
\usepackage[caption=false,font=normalsize,labelfont=sf,textfont=sf]{subfig}
\usepackage{textcomp}
\usepackage{stfloats}
\usepackage{url}
\usepackage{verbatim}
\usepackage{tikz}
\usepackage{graphicx}
\usepackage{cite}
\usepackage{color}
\usepackage{booktabs}
\usepackage{threeparttable}
\usepackage{multirow}
\begin{document}

\title{Spatiotemporal Regularized Tucker Decomposition Approach for Traffic Data Imputation}
\author{Wenwu Gong, Zhejun Huang, and Lili Yang

\thanks{W. Gong (ORCID: 0000-0002-8019-0582) is a Ph.D. student at the Department of Statistics and Data Science, Southern University of Science and Technology, Shenzhen, 518055, China.}

\thanks{Zhejun Huang is with the Department of Statistics and Data Science, Southern University of Science and Technology, Shenzhen, 518055, China.}

\thanks{Lili Yang is with the Department of Statistics and Data Science, Southern University of Science and Technology, Shenzhen, 518055, China.}
}

\markboth{Journal of \LaTeX\ Class Files,~Vol.~01, No.~1, Mar~2023}{Wenwu Gong \MakeLowercase{\textit{et al.}}: STRTD for Traffic Data Imputation}

% Remember, if you use this, you must call \IEEEpubidadjcol in the second

\maketitle

\begin{abstract}
In intelligent transportation systems, traffic data imputation, estimating the missing value from partially observed data is an inevitable and challenging task. Previous studies have not fully considered traffic data's multidimensionality and spatiotemporal correlations, but they are vital to traffic data recovery, especially for high-level missing scenarios. To address this problem, we propose a novel spatiotemporal regularized Tucker decomposition method. First, the traffic matrix is converted into a third-order tensor. Then, based on Tucker decomposition, the tensor is approximated by multiplying non-negative factor matrices with a sparse core tensor. Notably, we do not need to set the tensor rank or determine it through matrix nuclear-norm minimization or tensor rank minimization. The low rankness is characterized by the $l_1$-norm of the core tensor, while the manifold regularization and temporal constraint are employed to capture spatiotemporal correlations and further improve imputation performance. We use an alternating proximal gradient method with guaranteed convergence to address the proposed model. Numerical experiments show that our proposal outperforms matrix-based and tensor-based baselines on real-world spatiotemporal traffic datasets in various missing scenarios. 
\end{abstract}


\section{INTRODUCTION}
\IEEEPARstart{S}{patiotemporal} traffic data (STD) analysis is vital for road traffic control with intelligent transportation systems (ITS) development and application. For example, the road loop sensors record traffic state, including traffic speed, flow, and occupancy rate; the cars equipped with GPS (internet traffic data) record subjects’ movement from an origin to a destination, involving pair, time, and day modes. Both of them contain helpful information for traffic networks and route planning. Unfortunately, the missing data problem is inevitable due to communication malfunctions, transmission distortions, or adverse weather conditions \cite{CHEN2018patterns}. Consequently, the spatiotemporal traffic data imputation (STDI) is unavoidable and urgently required in ITS. 

Many imputation methods have been proposed to deal with the missing data problem, such as statistical-based methods \cite{imputeTS} and deep learning-based \cite{Thomas2021} methods. However, these methods either lack interpretability or have low accuracy. Due to spatiotemporal correlation and large-scale structure \cite{TAN201315, Chen20219548664}, low-rank tensor completion methods have been well developed. Among them, the low-rank tensor approximation (LRTA) model has proven to be highly effective in STDI theoretically and practically \cite{Kolda2009, Song2019}. Additionally, the LRTA model has successfully discovered interpretable traffic patterns, as reported by Chen et al. in 2018 \cite{CHEN2018patterns} and 2019 \cite{CHEN2019patterns}. The primary idea behind STDI is to characterize spatiotemporal correlations, as highlighted in the previous studies \cite{Rose2014, SaidSpatiotemporal}. Therefore, combining low rankness (long-term trends) and local correlations (short-term patterns) in traffic data is crucial in solving the STDI problem.

\subsection{Motivations}
This paper aims to capture traffic patterns from partially observed STDs via a factorization model and then use them to estimate the missing value accurately. Because low rankness provides long-term trends for the traffic data, the LRTA-based optimization model (referred to as the Tucker decomposition in this paper) with spatiotemporal regularization is used for the STDI problem. The motivations of our proposed model are three folds:

Firstly, the multidimensional array of STDs contains rich information. For example, traffic speed data in adjacent sensors show similar patterns and present temporal correlation properties \cite{Roughan2012, TASTWang}. The 2-dimensional traffic matrix imputation ignores the multidimensional nature and cannot deal with high missing rate scenarios \cite{CHEN2019patterns, LRHTWang2021}, especially for the structure missing scenario. Hence, reshaping the original traffic data into a high-order tensor to capture the traffic patterns is essential. 

Secondly, it is challenging to minimize the tensor rank. On the one hand, flattening the tensor into a matrix and minimizing the unfolding matrix nuclear norm is computationally time-consuming. On the other hand, rank determination remains the main challenge in using low-rank tensor decomposition models for STDI.

In addition, most previous tensor-based imputation methods only focused on the long-term trends and temporal patterns of STD, which made handling high-level and structured missing scenarios difficult. The Tucker decomposition model preserves the multidimensional nature of the STD and extracts the hidden patterns \cite{Xutao7460200} in a subspace. Thus, Tucker decomposition combined with spatiotemporal constraints in the subspace captures the traffic long term and reflects the spatial and temporal correlations for the STDI problem. 

\subsection{Contributions}
Though low-rank tensor completion is a hot topic for STDI, the problem is still open and needs to be better addressed. One of the main challenges is developing a low-rank Tucker model without a predefined rank that can accurately capture long-term trends. Considering the short-term patterns of the STD, another challenge remains to encode the spatiotemporal correlations and enhance the imputation performance.

This paper proposes an innovative enhanced low-rank Tucker decomposition model called Spatiotemporal Regularized Tucker Decomposition (STRTD) for the STDI problem. We summarize the main contributions as follows: 
\begin{enumerate}
	\item To better capture long-term traffic trends, we transform the matrix-based data into a 3rd-order tensor, which provides richer spatial and temporal information.
	\item We propose STRTD to characterize traffic patterns in STDs. The proposed model promotes long-term trends by contrasting the Tucker core tensor's sparsity and nonnegative factor matrices without a predefined rank. Additionally, STRTD employs manifold regularization and temporal constraint to characterize the short-term patterns and enhance the model performance;
	\item The proposed model is a highly nonconvex optimization problem with coupled variables. To decouple the objective function, we use the alternating proximal gradient (APG) method to obtain optimal solutions for STRTD.
	\item We verify the importance of spatiotemporal constraints in STRTD on two real-world STDs. A comprehensive comparative study with baselines is also conducted to demonstrate the effectiveness of STRTD. With the free-hyperparameters tuning, we demonstrate that STRTD performs better in real-world traffic imputation problems under different types of missing scenarios.
\end{enumerate}

We organize the rest of the paper as follows. Section \ref{sec: RW} discusses the related work on STDI. Section \ref{sec: notations} introduces the notations and defines the STDI problem. Section \ref{sec: STRTD} proposes the spatiotemporal constraints and model framework. In Section \ref{sec: Alg}, we present an algorithm that guarantees convergence. We evaluate the performance of our proposal on extensive experiments and compare them with some baseline approaches in Section \ref{sec: Num}. The last section concludes this paper and presents future work.

\section{RELATED WORK} 
\label{sec: RW}
Numerous time series imputation methods have been developed in the last two decades, especially for STDI \cite{WuTrac2020}. From the model-building perspective, these methods can be divided into machine- and low-rank learning-based. Because STD has spatial similarity and temporal variation characteristics, many studies have proven that the low rankness assumption combined with the spatiotemporal information method performs better than other existing methods for STDI \cite{SaidSpatiotemporal, Chen20219548664}. So, the low-rank tensor learning methods for STDI are discussed in detail in the following section.
\subsection{Low rankness}
The low-rank property, which depicts the inherent correlations in real-world datasets, is an essential and significant assumption in the completion problem. There are two categories of low-rank models to estimate the missing data problems: rank minimization and low-rank decomposition \cite{yu2016temporalr}. Regarding rank minimization, Candes et al. \cite{Candes2009} proposed the trace norm to estimate the missing matrix data precisely. Liu et al. \cite{Liu6138863} extended the matrix case to the tensor by proposing the nuclear norm for the image inpainting problem. Ran et al. \cite{RAN201654} applied the low-rank tensor nuclear norm minimization method to impute the spatiotemporal traffic flow. 

To avoid using the computationally expensive singular value decomposition (SVD) in unfolding matrix norm minimization, Tan et al. \cite{Huachun2014} proposed a Tucker decomposition model based on the truncated singular values of each factor matrix to exploit the low rankness in STD. Furthermore, Yokota et al. \cite{Yokota8578959} showed that the rank increment strategy is sufficient when a lower m-rank approximation initializes the tensor than its target m-rank in Tucker-based completion applications.

It can be seen that a significant difference between these two approximation methods is the way that they make decisions about low rankness. Compared with rank minimization and its relaxation, on the one hand, the low-rank decomposition model can preserve the tensor structure and avoid the high-cost unfolding matrix SVD \cite{PanLRSETD}. On the other hand, nuclear norm minimization cannot impose spatiotemporal constraints directly on traffic date \cite{TASTWang}. So, it is reasonable to believe that the low-rank decomposition model is more appropriate for the STDI task \cite{Rose2014, BGCPChen2019, CHEN2021Tubal}.

\subsection{Factorization Model}
Low-rank tensor decomposition, a high-order matrix factorization extension, has received increasing attention in spatiotemporal traffic data analysis. On the one hand, considering the sensory traffic matrix data, many papers applied the spatiotemporal Hankel operator to capture the low-rank structure by transforming the original incomplete matrix to a 4th-order tensor \cite{shi2020block, LRHTWang2021}. This transformer captures spatiotemporal information in a data-driven manner. However, it is time-consuming and parameter-sensitive. On the other hand, many papers used tensor decomposition models for STDI. For example, Tan et al. \cite{TAN201315, Huachun2014} proposed a Tucker decomposition-based model to estimate the missing traffic speed, guaranteeing accurate performance. Chen et al. \cite{BGCPChen2019} proposed a Bayesian augmented CP decomposition model for traffic data analysis, combining domain knowledge to enhance the imputation performance. Zhang et al. \cite{stTT2021} introduced tensor train (TT)-based models for imputing internet traffic data. However, estimating the exact rank of Tucker and CP decomposition in practice takes much work. Compared with the CP model, the Tucker decomposition can discover spatiotemporal patterns by interpreting the factor matrices \cite{CHEN2018patterns}. So, this paper focuses on the Tucker decomposition model for spatiotemporal traffic data imputation.

\subsection{Regularized Tucker Decomposition}
Tucker decomposition models can preserve the traffic data long-term information, which obtains low performance for STDI. Many studies have reported that using a low-rank Tucker decomposition model is insufficient for cases where the missing ratio is high \cite{Wu8421043, PanLRSETD, stTT2021}. To solve this problem, one of the most popular methods is to add regularization to the Tucker decomposition. Rose et al. \cite{Rose2014} proposed a unified low-rank tensor learning framework considering local similarity by constructing a Laplacian regularizer for multivariate data analysis.

Considering the spatiotemporal correlations in traffic data, the constraint-based methods, such as smoothness \cite{Wu8421043}, manifold regularization \cite{Roughan2012}, temporal regularization \cite{Chen20219548664}, and sparsity \cite{Goulart2017} have been well studied. For example, Zhang et al. \cite{ZHANG2019337} combined Tucker decomposition with the $l_2$ norm regularization of the factor matrix to detect and estimate traffic flows in the Automatic Number Plate Recognition system. Wang et al. \cite{Wang8708929} proposed a graph-regularized non-negative Tucker decomposition model to discover the interpretable traffic pattern from urban traffic flows. Goulart et al. \cite{Goulart201701} applied the orthogonal Tucker model with tensor core thresholding to adjust the core tensor size and introduced a feedback mechanism to enhance the traffic flow imputation performance. Besides, Pan et al. \cite{PanLRSETD} proposed a sparse enhanced Tucker decomposition model to exploit inherent long-term and short-term information in spatiotemporal traffic data imputation tasks. Most approaches require predefined Tucker ranks and are designed purely based on spatial or temporal correlation, resulting in low performance in data imputation, especially for structured missing scenarios.

To our knowledge, most papers still need to fully consider the long and short-term patterns simultaneously, i.e., low rankness and the spatiotemporal correlations. The proposed STRTD addresses these properties in a tensor object by reshaping the traffic matrix data in a 3rd-order spatiotemporal tensor form. Then, STRTD exploits the long-term traffic trends using a low-rank Tucker model without a predefined rank and captures the short-term patterns with given spatiotemporal priors, including manifold regularization and temporal constraint. Our experiment results demonstrate that STRTD performs better in two real-world STDs. 

\section{PRELIMINARIES}
\label{sec: notations}
We review some related concepts of Tucker decomposition as follows and present all notations used in this paper in Tab.~\ref{table1}. We refer \cite{Kolda2009} for more details of preliminaries. 
\subsection{Notations}
\begin{table}[!t]
\caption{Notations \label{table1}}
\centering
${
\begin{array}{r|l}
\hline \hline 
\mathcal{X}, \mathbf{U}, \alpha  & \begin{array}{l} \text{A tensor, matrix and real value, respectively.} \end{array}\\
\mathbb{R}_{+}^{I_1 \times I_2 \times \cdots \times I_N} & \begin{array}{l} \text {Set of N-th order nonnegative array.} \end{array} \\
\mathbf{U}\geq 0 & \begin{array}{l} \text {Nonnegative matrix } \mathbf{U} \text {, i.e., } u_{i j} \geq 0, \forall i, j. \end{array} \\
{\mathcal{P}_{+}(\mathbf{U})} & \begin{array}{l} \text{Operator yielding a nonnegative matrix of}  \\
{u_{i j}=} \max \left(u_{i j}, 0\right), \forall i, j.
\end{array} \\
{\mathcal{S}_{\mu}(x)} & \begin{array}{l} \text{Shrinkage operator with} \ \mu \text{ 
in component-wise.} 
\end{array} \\ 
\Omega, \bar{\Omega} & \begin{array}{l} \text {Observed index set and its complement.} \end{array} \\
\mathcal{X}_{\Omega} & \begin{array}{l} \text{Observed entries supported on the observed index.} \end{array} \\
\mathcal{H} & \begin{array}{l} \text {Tensorization operator.} \end{array} \\
\times_n & \begin{array}{l} \text {Mode-n product.} \end{array} \\
\otimes, \odot & \begin{array}{l} \text {Kronecker product and Hadamard product.} \end{array} \\
\left\| \ast \right\|_F & \begin{array}{l} \text {Frobenius norm.} \end{array} \\
\mathbf{X}_{(n)}  & \begin{array}{l} \text {Mode-n unfolding of tensor} \ \mathcal{X}. \end{array} \\
\operatorname{tr}  & \begin{array}{l} \text {Matrix trace operator}. \end{array} \\
\hline \hline
\end{array}
}$
\end{table}

A tensor is a multidimensional array where the order of the tensor is the number of dimensions, also called the mode. Throughout this paper, we use calligraphy font for tensors, such as ${\mathcal{X} \in \mathbb{R}^{I_1 \times I_2 \times \cdots \times I_N}}$, whose element is denoted as ${x_{i_{1}, i_{2}, \cdots, i_{n}}}$. The bold uppercase letters for matrices, such as ${\mathbf{U} \in \mathbb{R}^{I_1 \times I_2}}$, bold lowercase letters for vectors, such as ${\mathbf{a}\in \mathbb{R}^{I_1}}$, and lower case for scalars, such as ${\alpha, \beta}$. 

The Frobenius norm of a tensor is defined as 
$$\left\| \mathcal{X} \right\|^2_F = \sqrt{\sum_{i_{1}=1}^{I_{1}} \cdots \sum_{i_{n}=1}^{I_{N}} x_{i_{1} \ldots i_{n}}^{2}}.$$

We denote the mode-$n$ unfolding (i.e., matricization) of an $N$-order tensor $\mathcal{X}$ by $\mathbf{X}_{(n)} \in \mathbb{R}^{I_n\times \prod_{j \neq n} I_{j}}$.

The Tucker decomposition can be viewed as a higher-order principal component operator that minimizes the error in learning the projection on the subspace. Given a tensor ${\mathcal{X} \in \mathbb{R}^{I_1 \times I_2 \times \cdots \times I_N}}$, it can be decomposed into a core tensor $\mathcal{G}\in \mathbb{R}^{r_1 \times r_2 \times \cdots \times r_N}$ multiplying a matrix $\mathbf{U}_{n} \in \mathbb{R}^{I_n \times r_n}$ along each mode, i.e.,  ${\mathcal{X}=\mathcal{G} \times_{1}} {\mathbf{U}_{1} \cdots \times_{N} \mathbf{U}_{N}} = \mathcal{G} \times_{n=1}^{N} {\mathbf{U}_{n}}$. The core tensor $\mathcal{G}$ can be regarded as a compressed version of $\mathcal{X}$ if $r_1, r_2, \cdots r_N$ are significantly smaller than $I_1, I_2, \cdots I_N$. 

Based on the matrix Kronecker product $\otimes$, we can represent the Tucker decomposition by  $$\mathbf{X}_{(n)} =  \mathbf{U}_{n} \mathbf{G}_{(n)} \mathbf{V}_{n}^{\mathrm{T}} $$ 
where $\mathbf{V}_{n} = \left(\mathbf{U}_{N} \otimes \cdots \otimes \mathbf{U}_{n+1} \otimes \mathbf{U}_{n-1} \otimes \cdots \otimes \mathbf{U}_{1}\right)$ and the superscript `$\mathrm{T}$' represent matrix transpose. It is not difficult to verify that $\rm{vec}(\mathcal{X}) = \left(\mathbf{U}_{N} \otimes \cdots \otimes \mathbf{U}_{n} \otimes \cdots \otimes \mathbf{U}_{1}\right) \rm{vec}({\mathcal{G})} = \otimes_{n=N}^1 \mathbf{U}_{n} \rm{vec}({\mathcal{G})}$.

Finally, for a given tensor ${\mathcal{X} \in \mathbb{R}^{I_1 \times I_2 \times \cdots \times I_N}}$ and observed index set $\Omega$, we define ${\mathcal{X}_{\Omega}}$ as a projector that keeps the nonzero terms and leaves the other values as zero values, i.e., 
$${{\mathcal{X}_{\Omega}}:=\left\{\begin{array}{ll}
x_{i_{1} i_{2} \ldots i_{n}}, & \text { if }\left(i_{1}, i_{2}, \ldots, i_{n}\right) \in \Omega \\
0, & \text { otherwise. }
\end{array}\right.}$$

\subsection{Problem Definition}
\begin{figure}[!ht]
\centering
\includegraphics[width=3in]{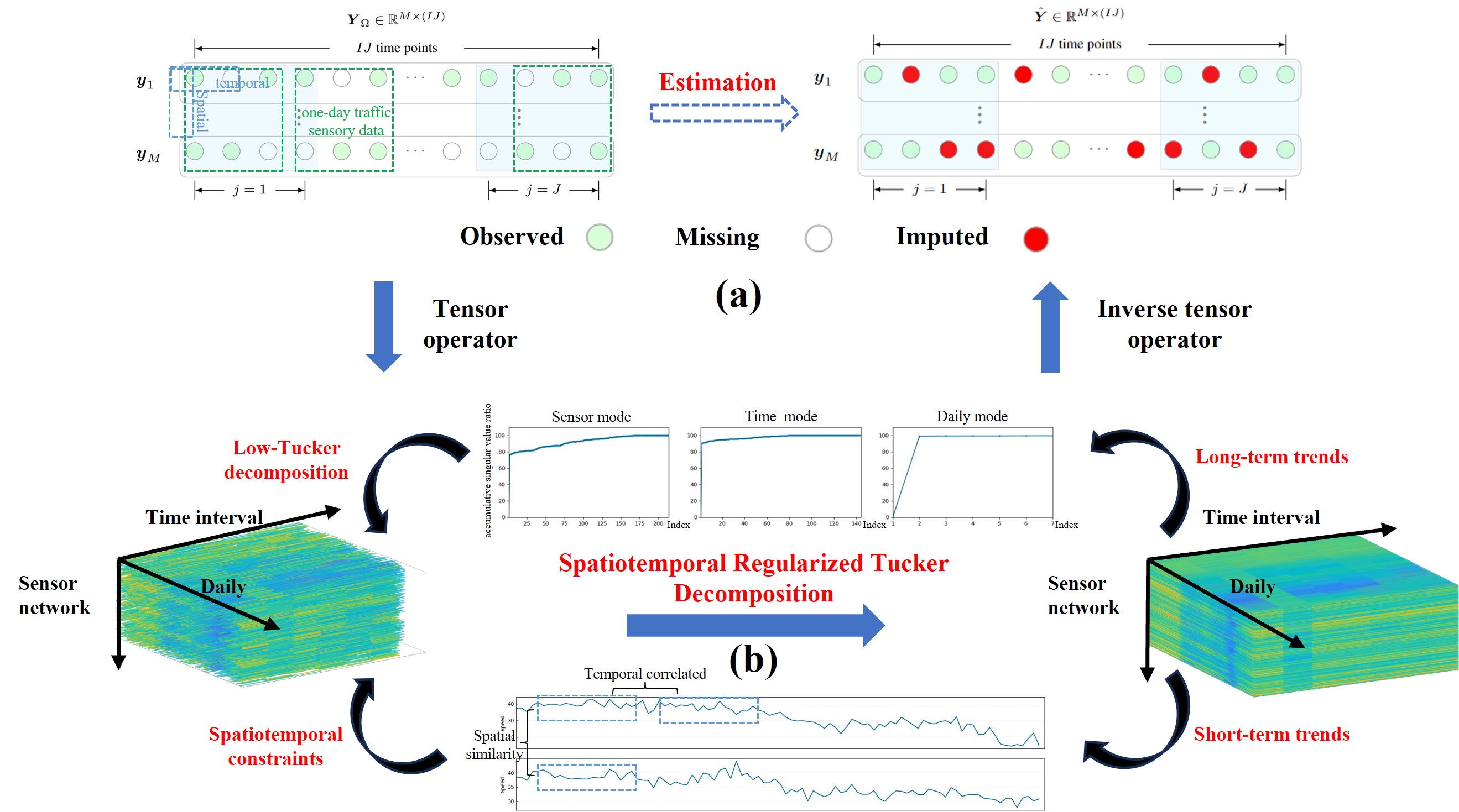}
\caption{The proposed STRTD framework for the STDI problem. (a) Matrix representation for STD. (b) Low-rank Tucker imputation based on the 3rd-order traffic tensor. }
\label{fig_0}
\end{figure}

STD is typically collected from $M$ sensors over $J$ days with $I$ time points. The missing multivariate time series is denoted as $\mathbf{Y}_{\Omega} \in \mathbb{R}^{M\times IJ}$ with the observed index set $\Omega$, as shown in Fig.~\ref{fig_0}(a). Chen et al. \cite{Chen20219548664} showed that a low-rank tensor can effectively capture long-term trends in STD and impute the traffic matrix. Furthermore, the STD tends to be similar along the nearby sensors and correlates at adjacent time points, reflecting short-term patterns \cite{Chen2022patterns}. So, this paper introduces the tensorization operator \cite{Yokota8578959} $\mathcal{H}$ to stack one-day traffic sensory data and reshape the STD into 3rd-order tensor. Then, an enhanced low-rank Tucker decomposition \cite{GongARTD} combined with the spatiotemporal constraints model (STRTD) is proposed to capture the long and short-term patterns in STD. Conversely, the inverse operator $\hat{\mathbf{Y}} =\mathcal{H}^{-1}(\hat{\mathcal{X}})$ converts the reconstructed tensor into the original traffic matrix and then imputes the missing values.  

Mathematically, we illustrate the proposed framework by minimizing the following objective function:
\begin{equation}
\begin{array}{cc}
     \underset{\mathcal{G}; \{\mathbf{U}_{n} \}}{\operatorname{minimize}}  \ \alpha \|\mathcal{G}\|_{1} + \beta \ g(\mathbf{U}_{n}) \\
     s.t.\ \mathcal{X} = \mathcal{G} \times_{n=1}^{N} {\mathbf{U}_{n}}, \ \mathbf{U}_{n} \geq 0, \quad \mathcal{X}_{\Omega} = \mathcal{H}(\mathbf{Y}_{\Omega}),
\end{array} \label{eq1}
\end{equation}
$g(\cdot)$ is the user-defined spatiotemporal constraint, and $\alpha, \beta $ are tradeoff parameters to compromise the low rankness and regularization role. Under different missing scenarios, we update $\mathcal{X}$ by the rule ~\eqref{Completion}
\begin{equation}
\hat{\mathcal{X}} = \mathcal{X}_{\Omega} + \{\hat{\mathcal{G}} \times_{1} {\hat{\mathbf{U}}_{1} \times \cdots \times_{N} \hat{\mathbf{U}}_{N}}\}_{\bar{\Omega}}. \label{Completion}
\end{equation} 
The main idea of our framework is to propose and study low-rank Tucker approximation for traffic tensor and then impute the traffic matrix by $\hat{\mathbf{Y}} =\mathcal{H}^{-1}(\hat{\mathcal{X}})$. 

\section{PROPOSED MODEL}
\label{sec: STRTD}
This section describes the formulation of STDI using spatial and temporal constraints in an enhanced low-rank Tucker decomposition model. The proposed method involves the sparsity of the Tucker core tensor, nonnegative factor matrices, manifold regularization, and temporal constraint.
\subsection{Spatiotemporal Constraints} 
As mentioned, STD often reflects short-term patterns along the spatial and temporal modes. On the one hand, the similarity between rows of the traffic matrix characterizes the spatial pattern, and the difference operator models the temporal variation \cite{Roughan2012}. On the other hand, the short-term patterns can be captured by using factor priors in the subspace under Tucker decomposition \cite{ChenSTDC, YuSBCD}. In this paper, we address the spatiotemporal correlations relying on the manifold regularization and temporal constraint matrix on factor matrices, which leads to better performance for the STDI problem.

\subsubsection{Manifold regularization} deals with non-linear data dimension reduction \cite{Xutao7460200}, which is used to search the geometric structure of the graph. Since the STD is in a low-dimensional spatial subspace, the similarity between the two sensors also exists in the spatial mode \cite{TASTWang}. We first select $p$ nearest neighbors in traffic sensors and use the kernel weighting to determine a similarity matrix, defined as \eqref{eq4}
\begin{equation}
  w_{i j} = e^{{-\left(\left\|{y}_{i}-{y}_{j}\right\|^{2}\right) / \sigma^{2}}}, \label{eq4}
\end{equation}
where $y_i$ and $y_j$ are the neighbor nodes along the spatial mode, $\sigma^2 = 1$ denotes the uniformly divergence. 

Given the matrix ${\mathbf{W} \in \mathbb{R}^{I_1\times I_1} \geq 0}$ for traffic spatial mode, we can use \eqref{eq2}, the manifold regularization term, to capture the spatial similarity in the subspace.
\begin{equation}
   {\sum_{i=1}^{I_1} \sum_{j=1}^{I_1} w_{i j}\left\|\mathbf{u}_{i}-\mathbf{u}_{j}\right\|_{2}^{2} =  \operatorname{tr}\left(\mathbf{U}^{\mathrm{T}} {\mathbf{L}} \mathbf{U}\right)}, \ \mathbf{L} = \mathbf{D}-\mathbf{W} \label{eq2}
\end{equation}
where $\mathbf{u}_{i}$ is the column vector of $\mathbf{U}^{\mathrm{T}}$ and ${\mathbf{D} \in \mathbb{R}^{I_1 \times I_1}}$ is a diagonal matrix with diagonal elements ${d_{i i}=}$ ${\sum_{j=1}^{I_1} w_{i j},}$ $i=1, \ldots, I_1$. Note that $\mathbf{L}$ is a Laplacian matrix designed by similarity matrix ${\mathbf{W}}$, which encodes the local similarity along the spatial mode \cite{Auxiliary2012}.

\subsubsection{Temporal constraint} it is to capture the correlations between adjacent time points in the time dimensional \cite{TASTWang}. Considering the non-stationary in the temporal dimension, the original traffic data is often correlated at adjacent time points. For adjacent $j-1$th and $j$th time points in traffic matrix $\mathbf{Y}$, we consider the Toeplitz operator $\mathbf{T}$ defined on the traffic tensor $\mathcal{X}$ to capture temporal variation, i.e., $\left\|\mathbf{Y}_{\cdot j} - \mathbf{Y}_{\cdot j-1}  \right\|^{2}_{F} = \left\|\mathcal{X} \times_{n} \mathbf{T}\right\|^{2}_{F}$. Note that
\begin{equation}
\begin{aligned}
\left\|\mathcal{X} \times_{n} \mathbf{T}\right\|^{2}_{F} & =\left\|\mathcal{G} \times_{1} {\mathbf{U}_{1}} \cdots \times_{n}\left({\mathbf{T}}{\mathbf{U}_{n}}\right) \times_{n+1} \cdots \times_{N} {\mathbf{U}_{N}}\right\|^{2}_{F} \\
& =\left\|\left({\mathbf{T}}{\mathbf{U}_{n}}\right)\left(\mathcal{G} \times_{p=1, p\neq n}^{N} {\mathbf{U}_{p}}\right)\right\|^{2}_{F} \\
& \leq\left\|{\mathbf{T}}{\mathbf{U}_{n}}\right\|^{2}_{F}\left\|\mathcal{G} \times_{p=1, p\neq n}^{N} {\mathbf{U}_{p}}\right\|^{2}_{F} \\
& \leq \text{const.}\left\|{\mathbf{T}}{\mathbf{U}_{n}}\right\|^{2}_{F}.
\end{aligned}
\end{equation}
Consequently, we use $\|\mathbf{T}\mathbf{U}\|_{F}^{2}$ to characterize the temporal correlation of traffic tensor in our proposal.

\subsection{Spatiotemporal Regularized Tucker Decomposition Model}
Let \( \mathcal{X}^0 \in \mathbb{R}^{I_{1} \times I_{2} \times \cdots \times I_{N}} \) (where N = 3 in our numeric experiments) be the missing traffic tensor, and \( \Omega \)  is the set of indexes corresponding to the observations. Based on the model \eqref{eq1} and the aforementioned spatiotemporal constraints, we consider the following optimization problem
\begin{equation}
  	\begin{array}{c}
  	\begin{aligned}
		\underset{\mathcal{G}; \{\mathbf{U}_{n}\}; \mathcal{X}}{\operatorname{minimize}} & \ \mathbb{F}(\mathcal{G}, \{\mathbf{U}_n\}, \mathcal{X}) \\ & \triangleq \{\frac{1}{2} \left\| \mathcal{X} - \mathcal{G} \times_{n=1}^{N} {\mathbf{U}_{n}}\right\|^2_F + \alpha\|\mathcal{G}\|_{1} +  \\
		& \sum_{n=1}^{K} \frac{\beta_{n}}{2} \operatorname{tr}\left(\mathbf{U}_{n}^{\mathrm{T}} \mathbf{L}_{n} \mathbf{U}_{n}\right) + \sum_{n=K+1}^{N} \frac{\beta_{n}}{2} \|\mathbf{T}_{n}\mathbf{U}_{n}\|_{F}^{2} \} \\ 
		s.t. & \ \mathbf{U}_{n} \in \mathbb{R}_{+}^{I_{n} \times I_{n}}, n=1, \ldots, N \ \text{and} \ \mathcal{X}_\Omega = \mathcal{X}^0_\Omega,
	\end{aligned}
	\end{array} \label{STRTD}
\end{equation}
where $ \alpha, \beta_{n} $ are positive penalty parameters, $K$ represents the numbers of spatial mode, $\mathbf{L}_n$ captures the spatial similarity, and $\mathbf{T}_{n}$ encodes the temporal variation. By imposing non-negativity constraints on the factor matrix, the Tucker core tensor becomes sparser \cite{GongARTD} and leads to a more intuitive explanation of traffic patterns \cite{Sinha_2022_WACV}. We name the model in \eqref{STRTD} as the Spatiotemporal Regularized Tucker Decomposition (STRTD) method, simultaneously exploiting the long and short-term characteristics of sensory traffic matrix data.

\noindent \textbf{\textit{Remark}:} The Tucker components' constraint assures that our proposal is well-defined. On the one hand, if all penalty parameters and nonnegative vanish, there are product combinations $\{\lambda_{n+1}\}$ such that $\{\lambda_1 \mathcal{G}, \lambda_2 \mathbf{U}_{1}, \cdots, \lambda_{n+1} \mathbf{U}_{n}\}$ does not change the value of \eqref{STRTD}. Hence, the low-rank Tucker approximation may not admit a solution. On the other hand, the spatiotemporal constraints imply the gradients of \eqref{STRTD} are Lipschitz continuous and have bounded Lipschitz constant under proximal linear operators (See \textit{\textbf{Proposition 1}} and \textit{\textbf{Proposition 2}}), which guarantee the solution set is nonempty.

\section{SOLVING STRTD MODEL}
\label{sec: Alg}
To solve the complicated optimization problems \eqref{STRTD}, we utilize the alternating proximal gradient (APG) method to update block variables by minimizing a surrogate function that dominates the original objective near the current iterate. Furthermore, we present the convergence and computational results for our proposed STRTD model.

\subsection{Optimization for the STRTD Model}
The STRTD is the regularized block multi-convex optimization problem \cite{Xu2012L2}, where we can use the prox-linear operator to solve that. The details are shown in Appendix A.

Firstly, we unfold \eqref{STRTD} in mode-$n$ for given tensor $\mathcal{X}$, then the factor matrices subproblems are given as the following three types.
\begin{itemize}
    \item Basic nonnegative matrix factorization.
    \begin{equation}
        \begin{array}{c}
  	    \begin{aligned}
        \underset{\mathbf{U}_{n} \geq 0}{\operatorname{minimize}} \ \ell({\mathbf{U}}_n) &=  \frac{1}{2}\left\|{\mathbf{X}_{(n)}}-\mathbf{U}_{n} {\mathbf{G}_{(n)}} {\mathbf{V}}^{\mathrm{T}}_{n}\right\|_{\mathrm{F}}^{2} 
        \end{aligned}
	    \end{array}\label{eq7}
    \end{equation}
    where $\mathbf{V}_{n} = \otimes_{p=N, p \neq n}^1 \mathbf{U}_{p}$. %${\mathbf{X}_{(n)}}$ and ${\mathbf{G}_{(n)}}$ are the mode-n unfolding of tensor $\mathcal{X}$ and $\mathcal{G}$, respectively. % $\mathbf{V}_{n} = \left(\mathbf{U}_{N} \otimes \cdots \otimes \mathbf{U}_{n+1} \otimes \mathbf{U}_{n-1} \otimes \cdots \otimes \mathbf{U}_{1}\right)$.
    \item Manifold regularization on factor matrix. 
    \begin{equation}
        \begin{array}{c}
  	    \begin{aligned}
        \underset{\mathbf{U}_{n} \geq 0}{\operatorname{minimize}} \ \ell({\mathbf{U}}_n) = &
       \frac{1}{2}\left\|{\mathbf{X}_{(n)}}-\mathbf{U}_{n} {\mathbf{G}_{(n)}} {\mathbf{V}}^{\mathrm{T}}_{n}\right\|_{\mathrm{F}}^{2} \\
        & + \frac{\beta_{n}}{2} \operatorname{tr}\left(\mathbf{U}_{n}^{\mathrm{T}} \mathbf{L}_{n} \mathbf{U}_{n}\right) 
        \end{aligned}
	    \end{array}\label{eq8}
    \end{equation}
    where ${\mathbf{L}_n = \mathbf{D}_n-\mathbf{W}_n}$ represents the Laplacian matrix.
    \item Temporal constraint on factor matrix. 
    \begin{equation}
      	\begin{array}{c}
  	    \begin{aligned}
        \underset{\mathbf{U}_{n} \geq 0}{\operatorname{minimize}} \ \ell({\mathbf{U}}_n) = & \frac{1}{2}\left\|{\mathbf{X}_{(n)}}-\mathbf{U}_{n} {\mathbf{G}_{(n)}} {\mathbf{V}}^{\mathrm{T}}_{n}\right\|_{\mathrm{F}}^{2} \\
        & \ + \frac{\beta_{n}}{2} \left\|\mathbf{T}_{n}\mathbf{U}_{n} \right\|_{\mathrm{F}}^{2} 
        \end{aligned}
	    \end{array} \label{eq9}
    \end{equation}
    where ${\mathbf{T}_{n}}$ is a self-defined temporal constraint matrix.
\end{itemize}

\noindent \textit{\textbf{Proposition 1:}} The objective function of subproblems \eqref{eq7} - \eqref{eq9} are differentiable and convex. Furthermore, the gradients $\nabla_{\mathbf{U}_n} \ell({\mathbf{U}}_n)$ are both Lipschitz continuous with bounded Lipschitz constant
\begin{equation*}
L_{\mathbf{U}_{n}}=\left\{\begin{array}{ll}
\left\|{\mathbf{G}_{(n)}} {\mathbf{V}}^{\mathrm{T}}_{n} {\mathbf{V}}\mathbf{G}_{(n)}^{\mathrm{T}}\right\|_{2} + \beta_{n} \left\|\mathbf{L}_{n} \right\|_{2}, & \text {Manifold} \\
\left\|{\mathbf{G}_{(n)}} {\mathbf{V}}^{\mathrm{T}}_{n} {\mathbf{V}}\mathbf{G}_{(n)}^{\mathrm{T}}\right\|_{2} + \beta_{n}  \left\| \mathbf{T}_{n}^{\mathrm{T}} \mathbf{T}_{n} \right\|_{2}, & \text {Temporal} \\
\left\|{\mathbf{G}_{(n)}} {\mathbf{V}}^{\mathrm{T}}_{n} {\mathbf{V}}\mathbf{G}_{(n)}^{\mathrm{T}}\right\|_{2}, & \text {otherwise }
\end{array}\right.
\end{equation*}

Then, \eqref{Matrix} presents the prox-linear operator to solve factor matrices subproblems. Appendix A contains the detailed proof of \textit{Proposition 1} and obtains updating rule \eqref{eq10}. 
\begin{equation}
    \hat{\mathbf{U}}_n = \underset{{\mathbf{U}_n} \geq 0}{\operatorname{argmin}} \ \left\langle\nabla_{\mathbf{U}_n} \ell(\tilde{\mathbf{U}}_n), \mathbf{U}_n-\tilde{\mathbf{U}}_n\right\rangle +\frac{L_{\mathbf{U}_n}}{2}\|\mathbf{U}_n-\tilde{\mathbf{U}}_n\|_{F}^{2}, \label{Matrix}
\end{equation}
where $\tilde{\mathbf{U}}_n$ denotes the extrapolated point.

Secondly, we update the subproblem $\mathcal{G}$ using the vectorization optimization problem \eqref{eq12}
\begin{equation}
\begin{array}{c}
  	\begin{aligned}
    \underset{\mathcal{G}}{\operatorname{minimize}} \ & \frac{1}{2}\left\|\rm{vec}(\mathcal{X}) - \otimes_{n=N}^1 \mathbf{U}_{n} \rm{vec}(\mathcal{G})\right\|_{\mathrm{F}}^{2} + \alpha\|\rm{vec}(\mathcal{G})\|_{1} \\
    & = f(\mathcal{G}) + \alpha\|\rm{vec}(\mathcal{G})\|_{1}.
    \end{aligned}
	\end{array}\label{eq12}
\end{equation}

\noindent \textit{\textbf{Proposition 2:}} The objective function of subproblem \eqref{eq12} is the
sum of two convex functions, and the gradient $\nabla_{\mathcal{G}} f(\mathcal{G})$ is Lipschitz continuous with the bounded Lipschitz constant $L_{\mathcal{G}} = \left\|\otimes_{n=N}^1 \mathbf{U}_{n}^{\top} \mathbf{U}_{n}\right\|_2 =\prod_{n=1}^{N}\left\|\mathbf{U}_{n}^{\mathrm{T}} \mathbf{U}_{n}\right\|_2$.

Based on \textit{Proposition 2}, we denote the core tensor prox-linear function as \eqref{Core} 
\begin{equation}
    \hat{\mathcal{G}} = \underset{\mathcal{G}}{\operatorname{argmin}}\left\langle\nabla_{\mathcal{G}} f(\tilde{\mathcal{G}}), \mathcal{G}-\tilde{\mathcal{G}}\right\rangle+\frac{L_\mathcal{G}}{2}\|\mathcal{G}-\tilde{\mathcal{G}}\|_{F}^{2}+\alpha\|\mathcal{G}\|_{1}, \label{Core}
\end{equation}
where $\tilde{\mathcal{G}}$ denotes the extrapolated point. Using the soft-thresholding operator \cite{Xu2015Tucker}, the core tensor updating rule is shown as \eqref{eq13}, and Appendix A presents the detailed proof.

Thirdly, considering the spatiotemporal priors are constrained on factor matrices $\{\mathbf{U}_{n}\}$, our proposed algorithm applies the order of $\mathcal{G}, \mathbf{U}_1, \mathbf{U}_2, \cdots, \mathbf{U}_N$ for algorithm design. Suppose the current iteration is $k$-th step, we update the core tensor $\mathcal{G}^k$ 
\begin{equation}
\begin{aligned}
    \mathcal{G}^{k+1} =  \mathcal{S}_{\frac{\alpha}{L_{\mathcal{G}}^{k}}}\left(\tilde{\mathcal{G}}^{k}-\frac{1}{L_{\mathcal{G}}^{k}} \nabla_{\mathcal{G}} f\left(\tilde{\mathcal{G}}^{k}\right)\right),  \label{eq13} 
\end{aligned}    
\end{equation}
where $\tilde{\mathcal{G}}^{k}$ is given by \eqref{GStep} and $\mathcal{S}_{\mu}(\mathcal{G}) $ is a soft-thresholding operator. Also, the factor matrices $\{\mathbf{U}_n^{k}\}$ is updated by
\begin{equation}
    \mathbf{U}_n^{k+1} = \mathcal{P}_{+}\left(\tilde{\mathbf{U}}_n^k - \frac{1}{L_{\mathbf{U}_n}^k} \nabla_{\mathbf{U}_n}\ell\left(\tilde{\mathbf{U}}_n^{k}\right)\right),  \label{eq10}
\end{equation}
where $\tilde{\mathbf{U}}_n^{k}$ is given by \eqref{UStep} and $\mathcal{P}_{+}(\mathbf{U})$ is a mapping function that projects the negative entries of $\mathbf{U}$ into zeros.

Technically, we propose an initial strategy where the $\{\mathbf{U}_n\}$ is generated randomly and then processed by normalization. By doing these, we conclude that it can reduce the low-rank approximation errors. Furthermore, we speed up Algorithm \ref{alg1} with \eqref{GStep} and \eqref{UStep}, which is updated by a parameterized iterative shrinkage-thresholding scheme \cite{FISTA2022}, shown in \eqref{Step}. 
\begin{equation}
\tilde{\mathcal{G}}^{k} ={\mathcal{G}^{k}}+\omega_{k}\left(\mathcal{G}^{k}-\mathcal{G}^{k-1}\right), \ \text{for} \ k \geq 1  \label{GStep}.
\end{equation}
\begin{equation}
\tilde{\mathbf{U}}_{n}^{k} ={\mathbf{U}_{n}^{k}}+\omega_{k}\left({\mathbf{U}_{n}^{k}}-{\mathbf{U}}_{n}^{k-1}\right), \ \text{for} \ k \geq 1  \label{UStep}.
\end{equation}
\begin{equation}
t^{k} =\frac{0.8+\sqrt{4 (t^{k-1})^{2}+0.8}}{2}, \quad \omega_{k} = \frac{t^{k-1}-1}{t^{k}}, 
\  t^{0} = 1.\label{Step}
\end{equation}

At the end of iteration $k$, we use the first-order feedback control rule \cite{GongARTD} to re-update tensor $\mathcal{X}^k$ when having $\{\mathbf{U}^k_n\}$ and $\mathcal{G}^k$
\begin{equation}
    {\mathcal{X}^{k+1}}_{\Omega} = {\mathcal{X}^{0}}_{\Omega}  + \gamma({\mathcal{X}^{k}}_{\Omega} - {\mathcal{Z}^{k}}_{\Omega}), \quad {\mathcal{X}^{k+1}}_{\bar{\Omega}} = {\mathcal{Z}^{k}}_{\bar{\Omega}}, \label{feedback}
\end{equation}
%\begin{equation}
    %{\mathcal{X}^{k+1}}_{\Omega} = {\mathcal{X}^{0}}_{\Omega}, \quad {\mathcal{X}^{k+1}}_{\bar{\Omega}} = {\mathcal{Z}^{k}}_{\bar{\Omega}}
%\end{equation}
where $ \mathcal{Z}^{k} = {\mathcal{G}^k} \times_{n=1}^{N}\mathbf{U}^k_{n}$, $\bar{\Omega}$ is the complement set of $\Omega$, and $0\leq \gamma \leq 1$ is a user defined hyper-parameter to control the correction. Furthermore, we ensure that the value of $\mathbb{F}\left(\mathcal{G}^{k}, \mathbf{U}_{k}\right)$ is decreasing before re-updating the $\tilde{\mathcal{G}}$, $\{\tilde{\mathbf{U}}_{n}\}$. If one of the following conditions is satisfied, we calculate the complete tensor $\hat{\mathcal{X}} = \mathcal{X}^0_{\Omega} + {\mathcal{Z}^{k}}_{\bar{\Omega}}$ as the imputed result.
\begin{equation}
\begin{array}{l}
\left\|\Omega \odot (\mathcal{Z}^{k}-\mathcal{X}^{0})\right\|_{F}\left\|\Omega \odot \mathcal{X}^{0}\right\|_{F}^{-1}<\text{tol}, \ \text{for some} \ k, \\
\text{or} \\
\frac{\left|\mathbb{F}_{\Omega}^{i}-\mathbb{F}_{\Omega}^{i+1}\right|}{1+\mathbb{F}_{\Omega}^{i}} \leq \text {tol}, \quad i = k,k+1,k+2, \label{eq11}
\end{array}
\end{equation} 
where $\mathbb{F}_{\Omega}^k$ denotes the objective function under observed index $\Omega$ at iteration $k$, and tol is a small specified positive value.

\begin{algorithm}[!ht]
	\caption{APG-based solver for the STRTD model}
	\label{alg1} 
	\begin{algorithmic}[1]
		\STATE \textbf{Input}: Missing traffic tensor $ {\mathcal{X}^{0} \in \mathbb{R}_{+}^{I_{1} \times I_{2} \times \cdots \times I_{N}}} $, $\Omega$ containing indices of observed entries, and the parameters $\alpha \geq 0 $, $\beta_{n} \geq 0 $, to = $1e^{-4}$, and $ K = 300 $.\\
		\STATE \textbf{Output}: Reconstructed tensor $\hat{\mathcal{X}}$. 
		\STATE \textbf{construct} positive semi-definite similarity matrix ${\mathbf{W}_{n}}$ and temporal constraint matrix ${\mathbf{T}_{n}}$;\\
		\STATE \textbf{initialize} $\mathcal{G}^0, \mathbf{U}^0_{n} \in \mathbb{R}_{+}^{I_n \times I_{n}} $ ($ 1 \leq n \leq N $);\\
		%\STATE $\mathcal{X}_{\Omega}$ = ${\mathcal{X}^0}_{\Omega}$, $\mathcal{X}_{\bar{\Omega}}$ = mean( $\mathcal{X}^0_{\bar{\Omega}}$);\\
		\FOR{$k=1$ to $K$}
            \STATE Optimize $ \mathcal{G} $ according to \eqref{eq13};
		\FOR{$n=1$ to $N$}
		\STATE Optimize $ \mathbf{U}_{n} $ using \eqref{eq10};
		\ENDFOR
		\STATE Update $ \mathcal{Z}^{k} $ using \eqref{feedback}; \\
		\STATE Whenever $\mathbb{F}\left(\mathcal{G}^{k}, \mathbf{U}_{k}\right) \textless  \ \mathbb{F}\left(\mathcal{G}^{k-1}, \mathbf{U}_{k-1}\right)$, we reupdate $\tilde{\mathcal{G}}$, $\{\tilde{\mathbf{U}}_{n}\}$ using \eqref{GStep} and \eqref{UStep} \textbf{until} stopping conditions \eqref{eq11} are satisfied.
		\ENDFOR
		\STATE \textbf{return} ${\hat{\mathcal{X}}_{\Omega}} = {\mathcal{X}^{0}}_{\Omega} $, \ $\hat{\mathcal{X}}_{\bar{\Omega}} = {\mathcal{Z}^{k}}_{\bar{\Omega}} $.
	\end{algorithmic}
\end{algorithm}

We conclude Algorithm \ref{alg1} as an APG-based updating procedure for the STRTD problem. As seen in Algorithm \ref{alg1}, each variable is updated with a closed-form solution, which improves the algorithm's efficiency.

\subsection{Convergence Analysis} 
Since the STRTD problem is nonconvex, obtaining the optimal global solution is difficult. It is shown in \cite{Xu2012L2} that regularized multi-convex optimization with cyclic block coordinate descent updating rule converges to a critical point. Since \eqref{STRTD} is a special case of the multi-convex optimization problem, we provide convergence property for the proposed algorithm as follows.
\begin{thm} 
    Let $\Theta^k = \{\{\mathbf{U}^k_n\},\mathcal{G}^k\}$ be the sequence generated by Algorithm \ref{alg1}, then we assure that $\Theta^k$ converges to a critical point $\hat{\Theta} = \{\{\hat{\mathbf{U}}_n\},\hat{\mathcal{G}}\} $.
\end{thm}
The above theorem ensures the feasibility of each solution produced by Algorithm 1. The proof of \textbf{Theorem 1} is based on the results of \textit{\textbf{Proposition 1}} and \textit{\textbf{Proposition 2}}. For simplicity, we give a proof framework here and present the details in Appendix B. Firstly, we establish a square summable result, i.e., ${\lim_{k \to \infty}\left(\Theta^{k}-\Theta^{k-1}\right) = 0}$. Next, we can prove $\hat{\Theta}$ is a stationary point by verifying the first-order optimality conditions. Finally, the KL inequality of $\mathbb{F}$ guarantees that $\Theta^k$ converges to a critical point.
\subsection{Computational Complexity Analysis}
Throughout this section, we denote the input tensor as ${\mathcal{X} \in \mathbb{R}^{I_{1} \times \ldots \times I_{N}}}$ and the core tensor as ${\mathcal{G} \in \mathbb{R}^{I_{1} \times \ldots \times I_{N}}}$. Here are the results of the computational cost of core ``shrinkage'' and latent factor matrix updating rules. The detailed analysis is shown in Appendix C.

Considering the proposed Algorithm \ref{alg1}, gradient computing is the most time-consuming. Moreover, Lipschitz constants calculation is negligible since the components can be obtained during the gradients' computation. Assuming that the STRTD converges in the $K$ iterations, we can roughly summarize the per-iteration complexity time complexity of the STRTD algorithm as
\begin{equation}
\begin{aligned}
	 \mathcal{O} \left( \left(N + 1\right) \sum_{n=1}^{N}\left(\prod_{i=1}^{n} I_{i}\right)\left(\prod_{j=n}^{N} I_{j}\right)\right),
\end{aligned}
\end{equation}
where the per-iteration cost is relevant to the tensor sizes $\prod_{i=1}^{n} I_{i}$, the proposed algorithm is efficient theoretically \cite{Xu2015Tucker}.

\section{EXPERIMENTS}
\label{sec: Num}
In this section, we conduct experiments on two STDs to compare STRTD with baselines in different missing scenarios. All experiments are performed using MATLAB 2023a on Windows 10 64-bit operating system on a workstation equipped with an Intel(R) Xeon(R) W-2123 CPU with 3.60 GHz, 64 GB RAM. Note that our Matlab codes are available on request.

\subsection{Traffic Datasets}  
We use the following two STDs for our experiment and form them as 3rd-order tensors for traffic data imputation problems. 
\begin{itemize}
    \item (\textbf{G}): Guangzhou urban traffic speed dataset. The original data is of size 214 $\times$ 8784 in the form of a multivariate time series matrix. We select seven days for our model training and reshape it into 3-rd order tensor of size $214 \times 144 \times 7$, i.e., (sensors, time, day).
    \item (\textbf{A}): Internet traffic flow dataset in Abilene. Dataset \textbf{A} includes 11 OD pairs, recording traffic flow every 5 minutes from December 8, 2003, to December 14, 2003. We consider a 3rd-order tensor of size $121\times 288\times 7$, where the first dimension corresponds to 121 OD pairs, the second to the time interval, and the last to 7 days.
    % \item (\textbf{P}): Portland highway traffic volume dataset. This dataset covers 743 road links and records 15-minute granularity volume data by 1156 loop coil detectors. The original data is of size $1156 \times 2976$ in a multivariate time series matrix. We select 50 detectors within seven days and reshape it to a 3rd-order tensor of size $50 \times 41 \times 7 \times 666$.
\end{itemize}

\begin{figure}[!ht]
\centering
\includegraphics[width=3.5in]{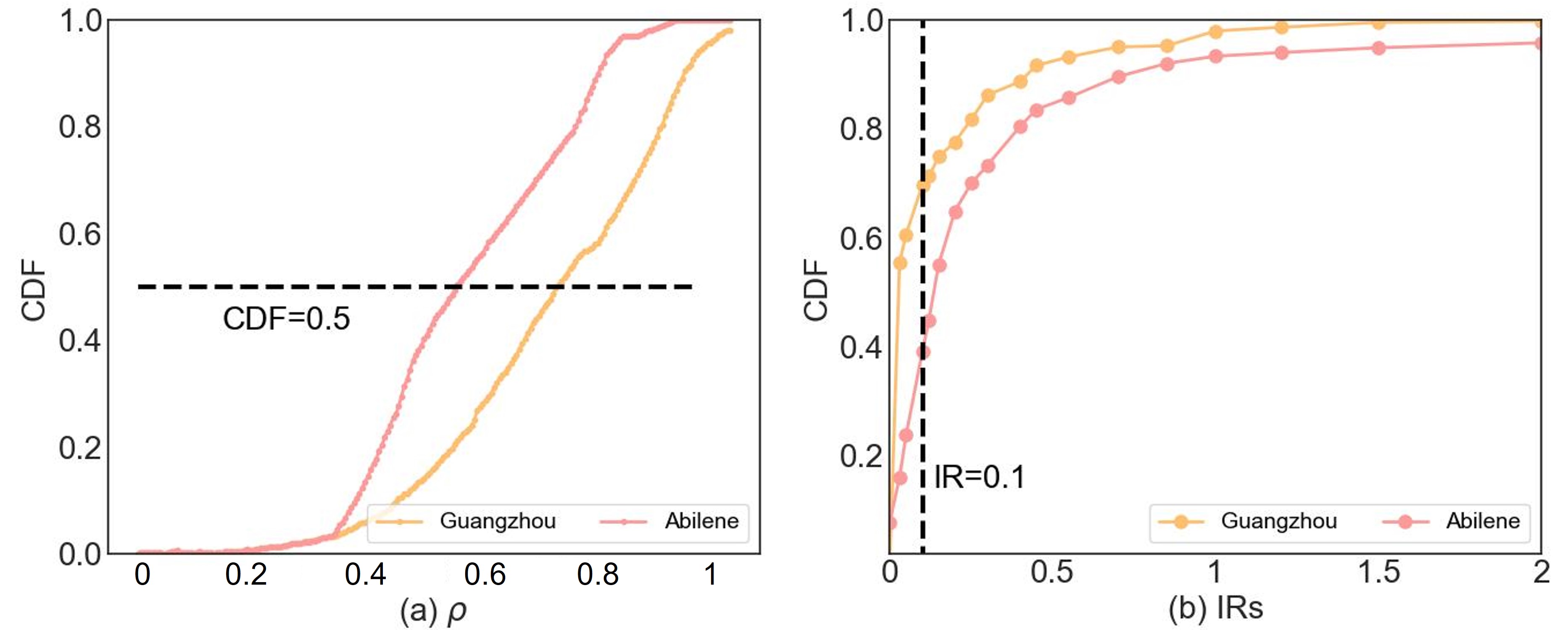}
\caption{Interpretation of the spatiotemporal characteristics in our STDs.}
\label{fig_1}
\end{figure}

To analyze these datasets' spatiotemporal characteristics, we first calculate the spatial correlations \cite{TAN201315} between various pairs of rows in traffic matrix $\mathbf{Y}$. Fig.~\ref{fig_1} (a) depicts the cumulative distribution function (CDF) of the correlation coefficient. It indicates that over 50\% of the sensors in two STDs exhibit strong correlations. This observation reveals that the sensor network in datasets \textbf{G} and \textbf{A} have strong spatial correlations. Fig.~\ref{fig_1}(b) shows the CDF of the traffic data with the increment rates (IRs) \cite{TASTWang}. More than 50\% of the data's IRs vary between 0.1 and 2, indicating temporal variations in the data. These results imply that the proposed spatiotemporal constraints are essential for our STDI problems.
\subsection{Experimental Settings}
\subsubsection{Missing scenario} For a thorough verification of the STRTD to STDI problem, we take into account three missing scenarios, i.e., random missing (RM), no-random missing (NM), and black-out missing (BM). Generally, RM means that missing data is uniformly distributed, and NM is conducted by randomly selecting sensors and discarding consecutive hours. At the same time, BM refers to all sensors not working for a certain period of time. According to the mechanisms, we mask the observed index set $\Omega$ and use the partial observations for the model training.
\subsubsection{Baseline models} For comparison, we select sixe state-of-the-art spatiotemporal traffic data imputation methods: stTT \cite{stTT2021}, LATC \cite{Chen20219548664}, LR-SETD \cite{PanLRSETD}, BGCP \cite{BGCPChen2019}, tSVD \cite{tSVD2017} and TAS-LR \cite{TASTWang}, to demonstrate the robustness and efficiency of our proposal. The baselines are shown in Tab.~\ref{table2}, in which the TAS-LR is a matrix-based approach, LATC is the matricization method, and others are the tensor decomposition method. 
\begin{table}[!ht]
\caption{Comparison of baseline models\label{table2}}
\centering
\begin{tabular}{ccccc}
\toprule
\multirow{2}{*}{\textbf{Baselines}} & \multicolumn{3}{c}{\textbf{Spatiotemporal constraints}} & \multirow{2}{*}{\textbf{Structures}} \\
& Low rankness & Spatial & Temporal \\
\midrule
STRTD & \checkmark & \checkmark & \checkmark & 3rd tensor\\
stTT \cite{stTT2021} & \checkmark & \checkmark & \checkmark & 3rd tensor \\
LATC \cite{Chen20219548664} & \checkmark & & \checkmark & 3rd tensor \\
LR-SETD \cite{PanLRSETD} & \checkmark & & \checkmark & 3rd tensor \\
BGCP \cite{BGCPChen2019} & \checkmark & & \checkmark & 3rd tensor \\
% LSTC \cite{CHEN2021Tubal} & \checkmark & & \checkmark & 3rd tensor \\
tSVD \cite{tSVD2017} & \checkmark & &  & 3rd tensor \\
% STH-LRTC \cite{LRHTWang2021} & \checkmark &  &  & 4th tensor \\
TAS-LR \cite{TASTWang} & \checkmark & \checkmark &  \checkmark & Matrix \\
% SRMF \cite{Roughan2012} & \checkmark & \checkmark & \checkmark & Matrix \\
\bottomrule
\end{tabular}
\begin{threeparttable}
\checkmark denotes the mentioned method has considered the constraint.
\end{threeparttable}
\end{table}
\subsubsection{Model performance} To measure the imputation performance, we adopt two criteria, including mean absolute percentage error (MAPE) and normalized mean absolute error (NMAE):
\begin{equation}
\begin{aligned}
    \mathrm{MAPE} &=\frac{1}{n} \sum_{i=1}^{n}\left|\frac{y_{i}-\hat{y}_{i}}{y_{i}}\right| \times 100, \\
    \mathrm{NMAE} &= \frac{\sum_{i=1}^{n} \left|y_{i}-\hat{y}_{i}\right|}{\sum_{i=1}^{n} \left|{y_{i}}\right|}
\end{aligned}
\end{equation}
where ${y_{i}}$ and ${\hat{y}_{i}}$ are actual values and imputed values, respectively.

\subsection{Implementation Details} 
\begin{figure}[htbp]
\centering
\includegraphics[width=3.5in]{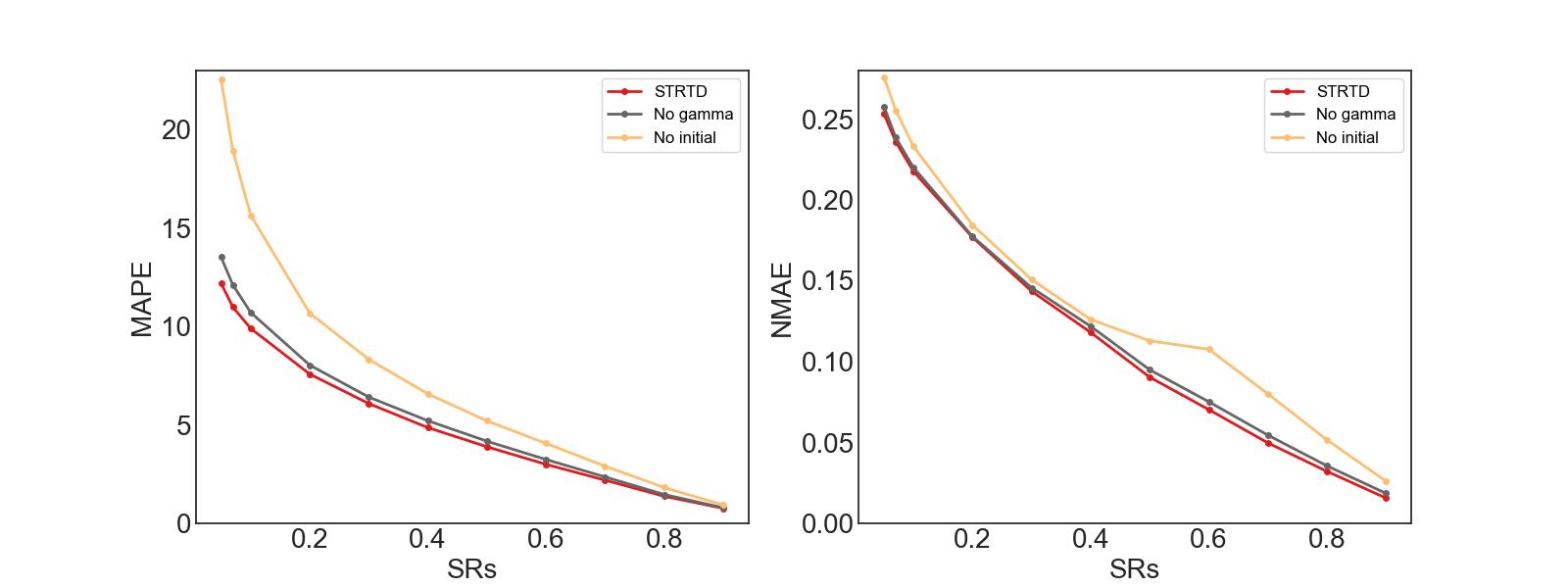}
\caption{Model performance over different strategies for \textbf{G} (left) and \textbf{A} (right), respectively.}	
\label{fig_2}
\end{figure}

\textit{Parameters setting:} Two parameters $\alpha$ and $\beta_{n}$ need to be tuned in our STRTD model. Hyperparameter $\alpha$ adjusts the strength of the sparsity term, i.e., the low-rank tensor approximation, and $\beta_{n}$ characterizes spatiotemporal regularization. In all our experiments, we easily set the core tensor size to be the same as the traffic tensor and set $\alpha = 1$, which does not need to predefine the Tucker ranks. We calculate the maximum SVD value of spatiotemporal constraint matrices to deliver $\beta_{n}$, i.e., $\beta_{n} = \frac{1}{2*0.1*\sigma (\mathbf{L} \ or \ \mathbf{T}\mathbf{T}^{\mathrm{T}})}$. Additionally, we evaluate the performance of different strategies to varying sample ratios (SRs) under RM scenarios, with SRs ranging from 0.9 to 0.1, 0.07, and 0.05. Fig. ~\ref{fig_2} shows that setting the parameter $\gamma$ to 0.2 reduces imputation error in high-level missing scenarios. In addition, the proposed initialization strategy reduces low-rank approximation errors. For better model comparison, the termination condition for all experiments is set to \eqref{eq11}, where tol = $10^{-4}$ and the maximum number of iterations is 300. Furthermore, the parameters of baselines are optimally assigned or automatically chosen as described in the reference papers.

\begin{figure*}[htbp]
\centering
\includegraphics[width=7.2in]{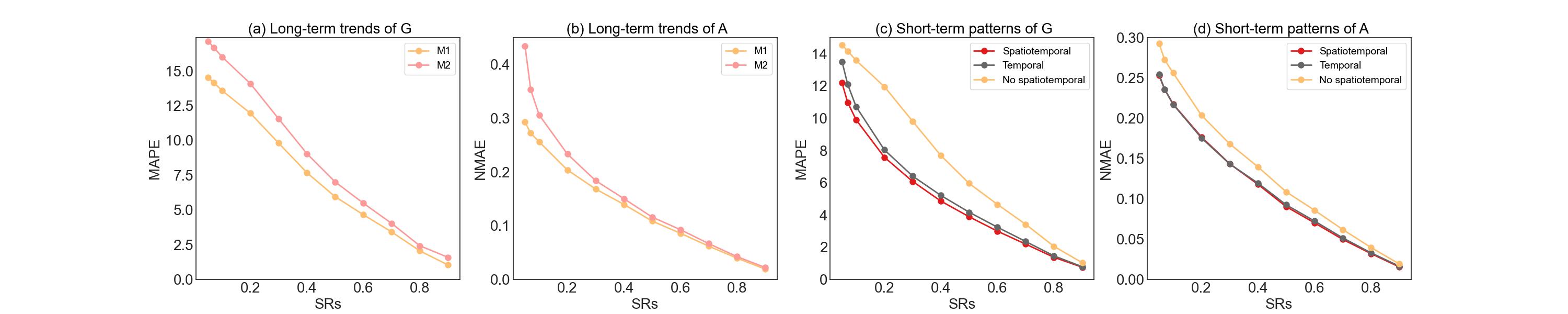}
\caption{Results of the ablation studies. (a)-(b) Interpretation of the multidimensionality of dataset \textbf{G} and \textbf{A}. (c)-(d) Illustration of the influence of spatiotemporal constraints for \textbf{G} and \textbf{A}, respectively.}	
\label{fig_3}
\end{figure*}

\textit{Ablation studies:} To illustrate STDs' long and short-term patterns, we first discuss the tensor structure in our STRTD model under different RM ratios. In our analysis, we denote the mentioned 3rd-order traffic tensor as \textbf{M1}. Following the method proposed in \cite{LRHTWang2021}, we reshape \textbf{G} into a $10 \times 205 \times 7 \times 1002$ tensor and \textbf{A} into a $11 \times 11 \times 288 \times 7$ tensor, represented by \textbf{M2}. Fig.~\ref{fig_3} (a)-(b) shows the model performance; it can be seen that the 3rd-order tensor structure covers richer spatial and temporal information. To further verify the validity of the spatiotemporal regularizations, we discuss the effect of the spatial and temporal constraints of STRTD. Fig.~\ref{fig_3}(c)-(d) compares the influence of spatiotemporal constraints for datasets \textbf{G} and \textbf{A}, respectively. We can observe that spatial and temporal regularizations enhance the traffic data imputation performance. However, compared with the spatial constraint, the temporal constraint plays a more important role.

\begin{figure}[!htbp]
\centering
\includegraphics[width=3.5in]{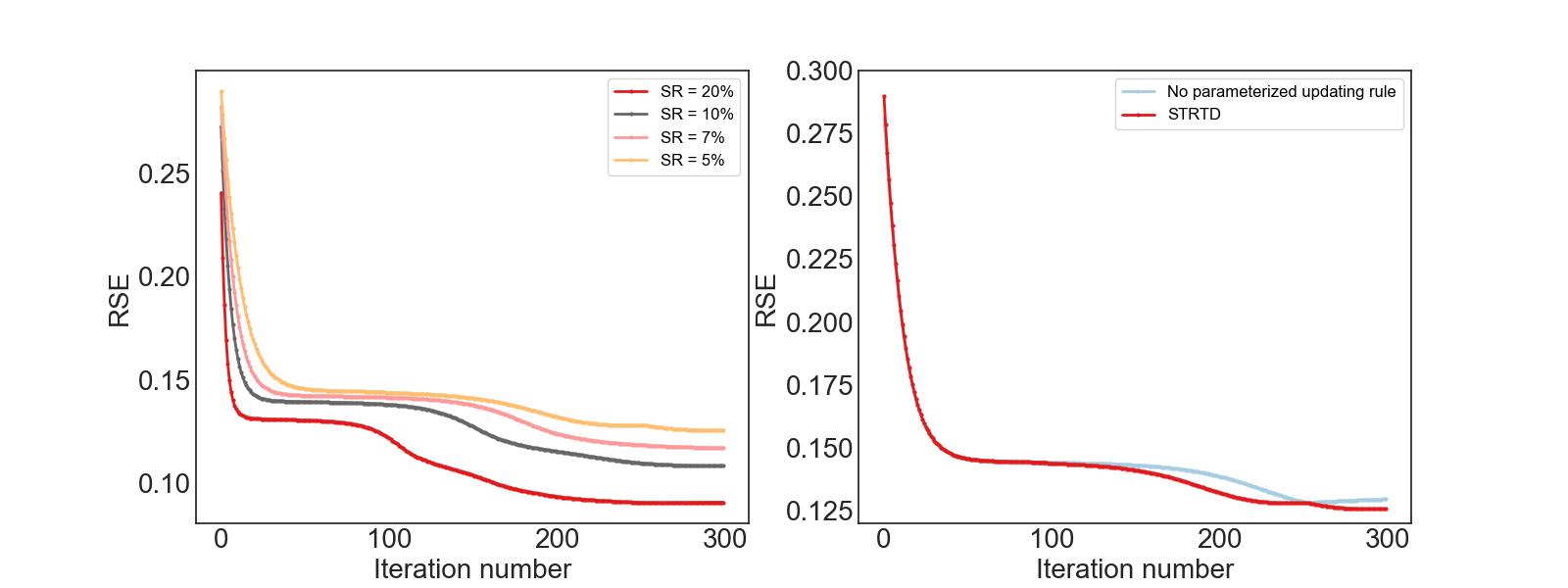}
\caption{The curves of the RSE values relative to the iterations under the
RM scenarios on \textbf{G} dataset for different SRs.}	
\label{fig_4}
\end{figure}

\textit{Convergence behaviors:} We have theoretically proven that the sequences generated by \textbf{Algorithm 1} converge to a stationary point in \textbf{Theorem 1}. Here, we show the numerical convergence of the proposed algorithm. Fig.~\ref{fig_4} (left) shows the curves of the relative square error (RSE) values versus the iteration number of the proposed STRTD on the \textbf{G} dataset. Remark that the RSE keeps decreasing as the iteration number increases, and the values stabilize after only about 200 iterations, which implies the proposed algorithm's numerical convergence. Furthermore, we test the RSE values of RM under SR = 0.05\%, Fig. ~\ref{fig_4} (right) shows that the parameterized updating rule can speed up the convergence of the proposed algorithm. 

\subsection{Results}
This section will compare the STRTD method with other baselines mentioned in Tab.~\ref{table3}.
\subsubsection{Overall performance among baseline models} As mentioned above, the ablation studies show that spatial and temporal constraints can improve the imputation performance for the STDI problem. To show the superiority of the STRTD, Tab.~\ref{table3} shows the overall imputation performance of baseline models on the datasets \textbf{G} and \textbf{A} under various missing scenarios. The best error indicator values are bolded. From these quantitative comparisons, the low-rank tensor imputation methods outperform matrix-based ones. In addition, the STRTD can impute the STDs with fewer observed data more accurately. Specifically speaking, the proposed method achieves the lowest MAPE and NMAE values. Compared with other baselines, we can see that the STRTD model performs better for each RM case. Reconstructing the NM and BM scenarios is more challenging than the RM scenarios, but the proposed method consistently performs well. This is because SRTD can utilize short-term traffic patterns in addition to long-term trends, confirming the advantages of combining low-rankness and spatiotemporal constraints.
\begin{table*}[!ht]
\caption{Performance comparison of STRTD and other baselines for RM, NM, and BM scenarios \label{table3}}
\centering
\begin{tabular}{cl|llllllll}
\toprule
{\textbf{Data}} & \multicolumn{1}{c|}{\textbf{Missing scenario}} & {\textbf{STRTD}} & {\textbf{tSVD}}& {\textbf{LATC}} & {\textbf{LR-SETD}} & {\textbf{BGCP}} & {\textbf{stTT}} &{\textbf{TAS-LR}}  \\ 
\midrule
\multirow{17}{*} & RM-30\% & {\textbf{2.19}} & 2.23 & 5.94 & 5.32 & 6.91 & 10.96 & 10.12 \\
& RM-70\% & {\textbf{6.08}} & 6.58 & 6.93 & 6.89 & 7.88 & 10.95 & 11.62 \\
& RM-90\% & {\textbf{9.90}} & 11.17 & 10.05 & 10.44 & 10.35 & 11.28 & 15.11 \\
\textbf{G} & RM-95\% & {\textbf{12.19}} & 13.55 & 12.60 & 16.22 & 12.25 & 12.84 & 17.69 \\
(MAPE) & NM-30\% & {\textbf{10.81}} & 12.93 & 74.98 & 13.48 & 15.62 &  11.88 & 12.31\\
& NM-70\% & {\textbf{12.48}} & 50.19 & 88.01 & 21.24 & 27.31 & 15.15 & 19.14\\
& NM-90\% & {\textbf{18.77}} & 85.51 & 87.66 & 57.33 & 32.73 & 21.97 & 52.01\\
& BM-30\% & {\textbf{13.56}} & 52.01 & 45.66 & 28.02 & 40.35 & 34.31 & 32.65 \\
\midrule
& RM-30\% & {\textbf{0.0497}} & 0.0501 & 0.1229 & 0.1159 & 0.1196 & 0.2120 & 0.3092 \\
& RM-70\% & {\textbf{0.1435}} & 0.1753 & 0.1488 & 0.1935 & 0.1527 & 0.2251 & 0.3178 \\
& RM-90\% & {\textbf{0.2175}} & 0.2274 & 0.2328 & 0.2356 & 0.2361 & 0.3764 & 0.3407 \\
\textbf{A} & RM-95\% & {\textbf{0.2532}} & 0.2752 & 0.4809 & 0.2601 & 0.3636 & 0.5124 & 0.3576 \\
(NMAE) & NM-30\% & {\textbf{0.2777}} & - & - & 0.6093 & - & 0.2869 & 0.3214\\
& NM-70\% & 0.4067 & - & - & 0.7401 & - & \textbf{0.3725} & 0.5791\\
& NM-90\% & {\textbf{0.7241}} & - & - & 0.8013 & - & 0.7303 & 0.8176\\
& BM-30\% & 0.4418 & - & - & \textbf{0.2509} & - & 0.3011 & 0.8417\\
\midrule
\multicolumn{2}{c|}{Time (Seconds)} & 31 & 17 & 140 & 35 & 1922 & \textbf{14} & 75 \\
\bottomrule      
\end{tabular}
\begin{threeparttable}
The best results are highlighted in bold fonts, and - denotes that the algorithm is not applicable.
\end{threeparttable}
\end{table*}

\begin{figure*}[htbp]
\centering
\includegraphics[width=7in]{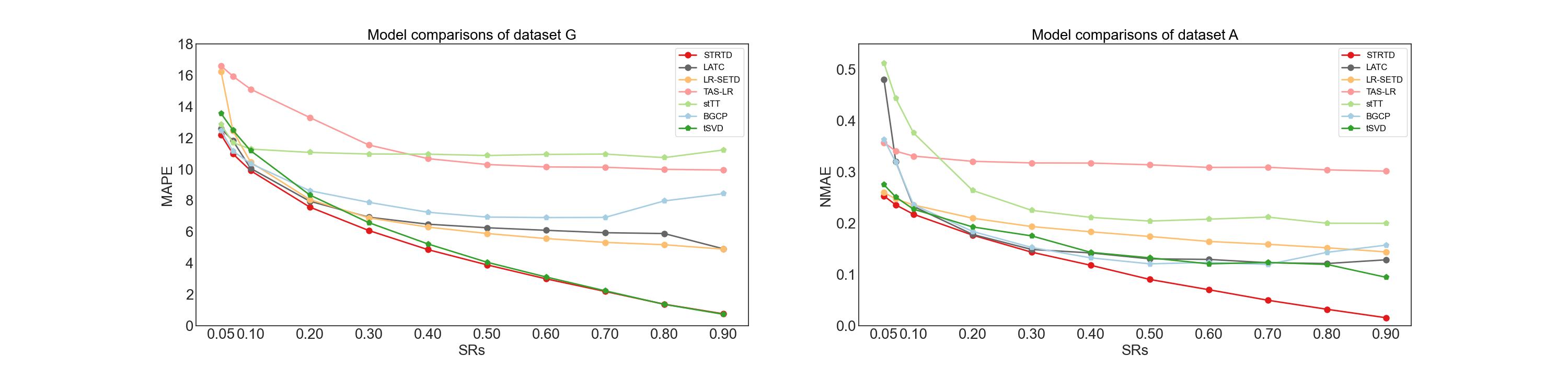}
\caption{MAPE and NMAE values for different sample ratios under RM scenarios for datasets \textbf{G} (left) and \textbf{A} (right), respectively. }
\label{fig_5}
\end{figure*}

To further compare the model performance of our proposal STRTD, we calculate the MAPE values for the \textbf{G} and NMAE values for the \textbf{A} under different RM scenarios (SR changes from 0.90 to 0.05) in Fig.~\ref{fig_5}. It is easy to see that STRTD has the lowest value, even for the extremely missing. Especially when the missing rate is 95\%, imputing with MAPE and NMAE values results in improvements higher than 3\%. 

\subsubsection{Imputation examples with different missing scenarios} 
\begin{figure*}[!htbp]
\centering
\includegraphics[width=6in]{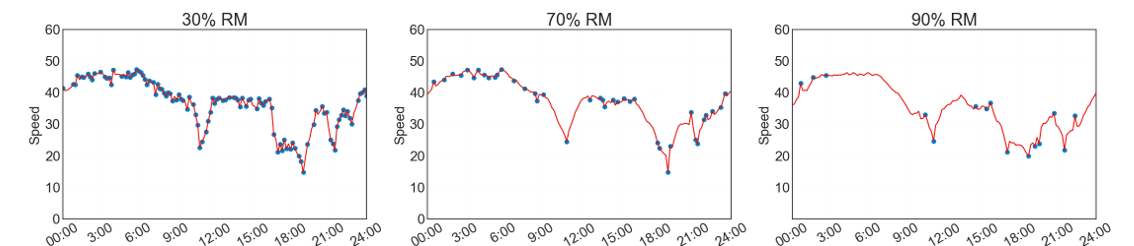}
\caption{Results of RM scenario on \textbf{G} dataset. This example corresponds to the 81st sensor and the 4th day of the dataset. Purple dots indicate the partially observed data, and red curves indicate the imputed values. }
\label{fig_8}
\end{figure*}

\begin{figure*}[!ht]
\centering
\includegraphics[width=6in]{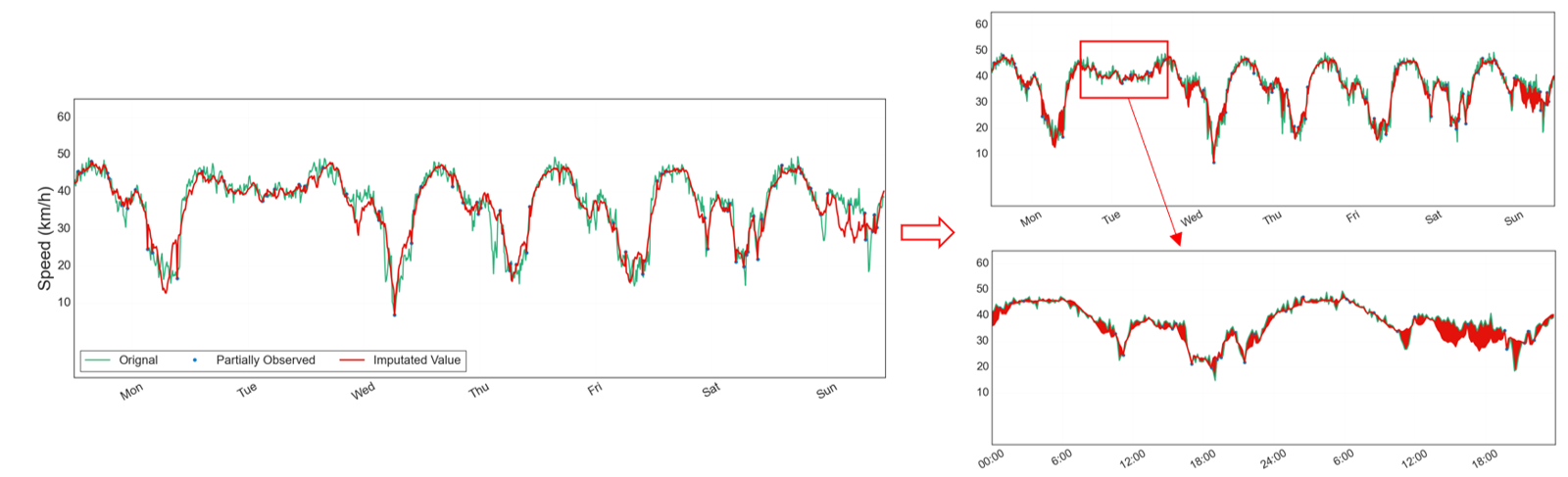}
\caption{Imputed values by STRTD on \textbf{G} dataset under RM scenario with 90\% missing. Note that the red area (residual area) is only used to express the estimation performance, which does not represent the cumulative residual.}
\label{fig_9}
\end{figure*}

\begin{figure*}[!htbp]
\centering
\includegraphics[width=6in]{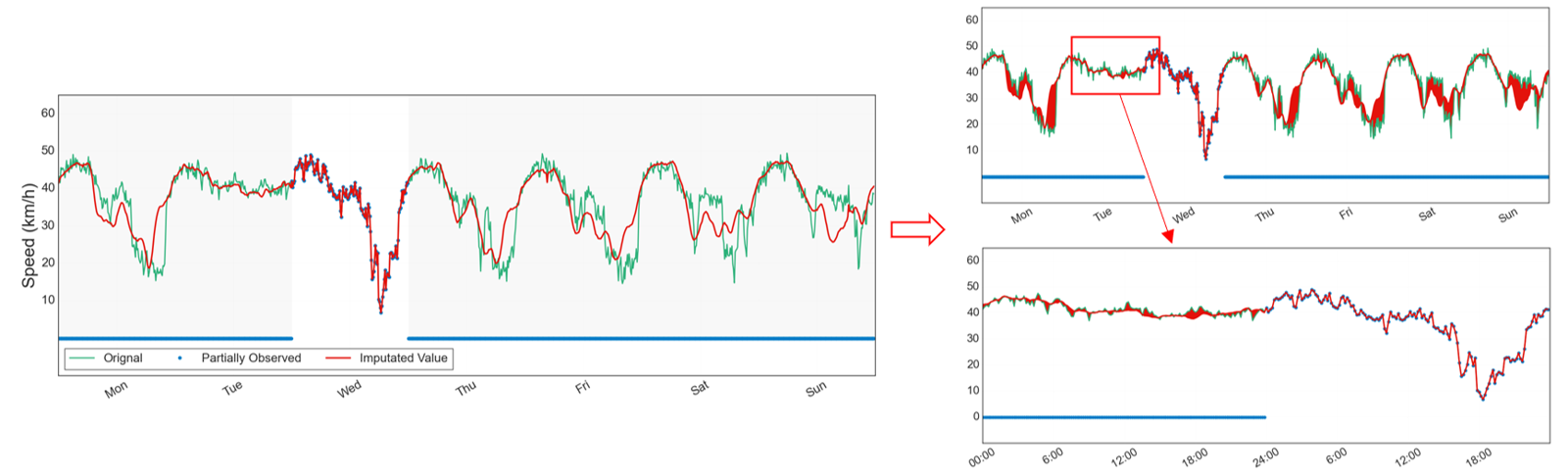}
\caption{Imputed values by STRTD on \textbf{G} dataset under NM scenario with 70\% missing. The gray rectangles indicate the missing area.}
\label{fig_10}
\end{figure*}

\begin{figure*}[!htbp]
\centering
\includegraphics[width=6in]{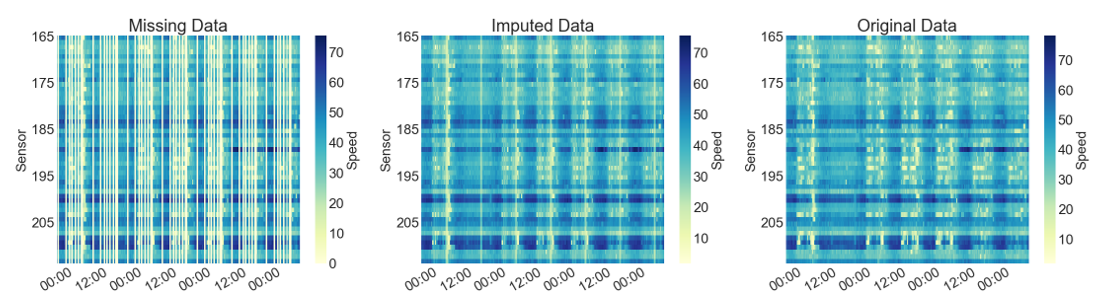}
\caption{The visualization of STRTD on \textbf{G} dataset under BM scenario with 30\% missing. The middle heat map is our STRTD results.}
\label{fig_11}
\end{figure*}

Here, we show some STRTD imputation examples with different missing scenarios on the Guangzhou (\textbf{G}) dataset. For the RM scenario, Fig.~\ref{fig_8} shows the same signal trends under different SRs (see the number of purple dots), indicating that the STRTD can accurately impute partial observations. Also, the residuals in Fig.~\ref{fig_9} reveal that the STRTD can successfully reconstruct the STD precisely, even for the extreme case (i.e., 90\% RM). 
To further validate the superiority of our STRTD, we plot the structural missing scenarios (NM and BM) result in Fig.~\ref{fig_10} and Fig.~\ref{fig_11}. In this case, the STRTD can achieve accurate imputation and learn traffic trends from severe missing scenarios.

\section{CONCLUSION}
Spatiotemporal traffic data imputation (STDI) is an inevitable and challenging task in data-driven intelligent transportation systems (ITS). This paper treats the STDI as a low-rank Tucker decomposition problem. The proposed STRTD exploits the long-term trends using a low-rank Tucker model and captures the short-term patterns with manifold regularization and temporal constraint. Through extensive experiments on two real-world STDs, our results show that the proposed STRTD beats other baselines for STDI with different RM scenarios and performs well on NM and BM missing scenarios (see Tab.~\ref{table3} and Fig.~\ref{fig_5}).

There are three potential prospects for future work. First, our proposal ignores the exact rank and uses the sparse core tensor and nonnegative factor matrix terms to promote low rankness. A potential approach is to use another low-rank tensor measure, such as multiplying the factor matrix rank to encode the Tucker rank \cite{XieKBR2}. Second, the current framework suffers a high computational cost for large-scale matrix multiplication calculations. One can consider the fast Fourier transform to address this issue \cite{Tatsuya2022}. Third, in addition to STDI, we can use the STRTD for spatiotemporal traffic data forecasting even with the missing observations \cite{Chen2022Prediction}. Also, the proposed spatiotemporal traffic data modeling frameworks can be considered for urban traffic pattern discovery \cite{Chen2022patterns}.

\section*{Acknowledgments}
The authors would like to thank the authors who shared their code and data on websites. This research is partly supported by the Shenzhen Science and Technology Program (Grant No. ZDSYS20210623092007023, JCYJ20200109141218676, K22627501) and Guangdong Province Universities and Colleges Key Areas of Special Projects (Grant No. 2021222012).

{\appendices
\section{APG-based Algorithm for the STRTD}
We first denote the STRTD optimization problem as a class of regularized block multi-convex optimization problems:
$$\underset{\{\mathbf{x}_n\}}{\operatorname{minimize}} \ \ell\left(\{\mathbf{x}_n\}\right) + \sum_{n=1}^{N} \beta_{n} f_{n}\left(\mathbf{x}_{n}\right),$$
%$$\underset{\mathcal{G}; \{\mathbf{U}_n\} \geq0 }{\operatorname{minimize}} \ \ell\left(\mathcal{G},\{\mathbf{U}_n\}\right) +\alpha r(\mathcal{G}) + \sum_{n=1}^{N} \beta_{n} r_{n}\left(\mathbf{U}_{n}\right),$$
where $f_{n}\left(\mathbf{x}_{n}\right)$ is the given constraint. We consider the APG-based prox-linear operator to update every $\mathbf{x}_n$ by solving a relaxed subproblem with a separable quadratic objective:
\begin{equation}
    \begin{array}{c}
  	\begin{aligned}
    \mathbf{x}_{n}^{k}=\underset{\mathbf{x}_{n}}{\operatorname{argmin}} & \{ \left\langle\tilde{\mathbf{g}}^{k}, \mathbf{x}_{n}-\tilde{\mathbf{x}}_{n}^{k-1}\right\rangle+\frac{L^{k-1}_{\mathbf{x}_{n}}}{2}\left\|\mathbf{x}_{n}-\tilde{\mathbf{x}}_{n}^{k-1}\right\|_F^{2} \\
    &+\beta_{n} r_{n}\left(\mathbf{x}_{n}\right)\},
    \end{aligned}
	\end{array}\label{APG}
\end{equation}
where $\tilde{\mathbf{x}}_{n}^{k-1}$ denotes an extrapolated point and update through
\begin{equation}
\begin{aligned}
\tilde{\mathbf{x}}_{n}^{k-1} &={\mathbf{x}_{n}^{k-1}}+\omega_{k-1}\left({\mathbf{x}_{n}^{k-1}}-{\mathbf{x}}_{n}^{k-2}\right), \ \text{for} \ k \geq 1 \\
\omega_{k-1} &= \frac{t^{k-2}-1}{t^{k-1}}, \quad t^{k-1} =\frac{p+\sqrt{r (t^{k-2})^{2}+q}}{2},
\end{aligned} 
\end{equation}
where $p, q \in \left[0,1\right]$, $r \in \left[0,4\right]$, and $\tilde{\mathbf{g}}^k = \nabla_{\mathbf{x}_n} \ell \left(\tilde{\mathbf{x}}_{n}^{k-1}\right) $ is the partial gradient of objective function $\ell$. Guided by \cite{FISTA2022}, we set $p = q = 0.8, r = 4$ and the updating rule \eqref{APG} under these sequences have $\mathcal{O} \left(1/{k^2}\right)$ convergence rate.

Then, we provide a detailed proof of \rm{\textbf{Proposition 1}} and \rm{\textbf{Proposition 2}}, followed by an explanation of the closed-form updating rule.
\begin{proof}[\rm{\textbf{Proof of Proposition 1}}] Obviously, the Frobenius norm and matrix trace are differentiable functions. It remains to prove the convex property of $\ell$ and the Lipschitz continuous property of $\nabla_{\mathbf{U}_{n}} \ell$. Let $ \Phi = \frac{1}{2}\left\|\mathbf{X}_{(n)}-\mathbf{U}_{n} {\mathbf{G}_{(n)}} \mathbf{V}_{n}^{\mathrm{T}}\right\|_{\mathrm{F}}^{2} $ and $ g = \operatorname{tr}\left(\mathbf{U}_{n}^{\mathrm{T}} {\mathbf{L}_{n}} \mathbf{U}_{n}\right) \ \text{or} \ \left\|\mathbf{T}_{n} \mathbf{U}_{n} \right\|_{\mathrm{F}}^{2} $, we have the gradient of $ \ell(\mathbf{U}_{n}) = \Phi(\mathbf{U}_{n}) + \frac{\beta_{n}}{2} \Phi(\mathbf{U}_{n}) $
\begin{equation}
\nabla_{\mathbf{U}_{n}} \ell(\mathbf{U}_{n}) = {\mathbf{U}_{n}} {\mathbf{G}^{n}_{\mathbf{V}}} {\mathbf{G}^{n}_{\mathbf{V}}}^{\mathrm{T}} - \mathbf{X}_{(n)} {\mathbf{G}^{n}_{\mathbf{V}}}^{\mathrm{T}} + \nabla_{\mathbf{U}_{n}} g(\mathbf{U}_{n}). \label{A1}
\end{equation}
where ${\mathbf{G}^{n}_{\mathbf{V}}} = {\mathbf{G}_{(n)}} \mathbf{V}_{n}^{\mathrm{T}}$ and $\mathbf{V}_{n} = \left(\otimes_{p\neq n}^1 \mathbf{U}_{p}\right)$. 

On the one hand, the Hessian matrix of $\ell(\mathbf{U}_{n})$ is given by
\begin{equation}
\nabla_{\mathbf{U}_{n}}^2 \ell(\mathbf{U}_{n})=\left\{\begin{array}{ll}
\mathbf{G}^{n}_{\mathbf{V}} {\mathbf{G}^{n}_{\mathbf{V}}}^{\mathrm{T}} +  \beta_{n} \mathbf{L}_{n}, & \text {Manifold} \\
\mathbf{G}^{n}_{\mathbf{V}} {\mathbf{G}^{n}_{\mathbf{V}}}^{\mathrm{T}} +  \beta_{n} \mathbf{T}_{n}^{\mathrm{T}} \mathbf{T}_{n}, & \text {Temporal} \\
\mathbf{G}^{n}_{\mathbf{V}} {\mathbf{G}^{n}_{\mathbf{V}}}^{\mathrm{T}}, & \text {otherwise }
\end{array}\right. \label{A2}
\end{equation}
As we know, the functions $\mathbf{G}^{n}_{\mathbf{V}} {\mathbf{G}^{n}_{\mathbf{V}}}^{\mathrm{T}}$, $\mathbf{L}_{n}$ and $\mathbf{T}_{n}\mathbf{T}_{n}^{\mathrm{T}}$ are both positive semidefinite. According to the definition of a convex function, we know $\ell(\mathbf{U}_{n})$ is convex. 

On the other hand, we need the Lipschitz constant of $\nabla_{\mathbf{U}_{n}} \ell$. Since $\ell(\mathbf{U}_{n})$ is a linear combination of $ \Phi(\mathbf{U}_{n}) $ and $ g(\mathbf{U}_{n}) $, the Lipschitz constant of $\nabla_{\mathbf{U}_{n}} \ell$ can be calculated as a linear combination of the Lipschitz constants of the $\nabla_{\mathbf{U}_{n}} \Phi $ and ${\nabla_{\mathbf{U}_{n}} g}$. Such as, taken $ g(\mathbf{U}_{n}) = \frac{\beta_{n}}{2} \operatorname{tr}\left(\mathbf{U}_{n}^{\mathrm{T}} \mathbf{L}_{n} \mathbf{U}_{n}\right) $
\begin{equation}
\nabla_{\mathbf{U}_{n}} \ell(\mathbf{U}_{n}) = {\mathbf{U}_{n}} {\mathbf{G}^{n}_{\mathbf{V}}} {\mathbf{G}^{n}_{\mathbf{V}}}^{\mathrm{T}} - \mathbf{X}_{(n)} {\mathbf{G}^{n}_{\mathbf{V}}}^{\mathrm{T}} + \beta_{n} \mathbf{L}_{n} {\mathbf{U}_{n}}.
\end{equation}
For any two matrices ${\mathbf{U}_{n}^1}, {\mathbf{U}_{n}^2}$, we have 
\begin{equation}
	\begin{array}{l}
		\left\|\nabla_{\mathbf{U}_{n}} \ell(\mathbf{U}_{n}^1) - \nabla_{\mathbf{U}_{n}} \ell(\mathbf{U}_{n}^2) \right\|_{\mathrm{F}}^{2} \\ 
		= \left\|\left({\mathbf{U}_{n}^1} - {\mathbf{U}_{n}^2}\right) {\mathbf{G}^{n}_{\mathbf{V}}} {\mathbf{G}^{n}_{\mathbf{V}}}^{\mathrm{T}} - \beta_{n} \mathbf{L}_{n} \left({\mathbf{U}_{n}^1} - {\mathbf{U}_{n}^2}\right) \right\|_{\mathrm{F}}^{2} \\
		\leq \left\|\left({\mathbf{U}_{n}^1} - {\mathbf{U}_{n}^2}\right) {\mathbf{G}^{n}_{\mathbf{V}}} {\mathbf{G}^{n}_{\mathbf{V}}}^{\mathrm{T}} \right\|_{\mathrm{F}}^{2} + \left\|\beta_{n} \mathbf{L}_{n} \left({\mathbf{U}_{n}^1} - {\mathbf{U}_{n}^2}\right) \right\|_{\mathrm{F}}^{2}. \\
	\end{array} 
\end{equation}
So, we only need to calculate the Lipschitz constant of the composite gradient $\nabla_{\mathbf{U}_{n}} \ell$ separately. More specifically,
\begin{equation}
	\begin{array}{l}
		\quad \left\|\left({\mathbf{U}_{n}^1} - {\mathbf{U}_{n}^2}\right) {\mathbf{G}^{n}_{\mathbf{V}}} {\mathbf{G}^{n}_{\mathbf{V}}}^{\mathrm{T}} \right\|_{\mathrm{F}}^{2} \\
		= \operatorname{tr}\left({\mathbf{G}^{n}_{\mathbf{V}}} {\mathbf{G}^{n}_{\mathbf{V}}}^{\mathrm{T}} \left({\mathbf{U}_{n}^1} - {\mathbf{U}_{n}^2}\right)^{\mathrm{T}}\left({\mathbf{U}_{n}^1} - {\mathbf{U}_{n}^2}\right){\mathbf{G}^{n}_{\mathbf{V}}} {\mathbf{G}^{n}_{\mathbf{V}}}^{\mathrm{T}}\right) \\
		%= \operatorname{tr}\left({S}_{orth}^{\mathrm{T}}\left({\mathbf{U}_{n}^1} - {\mathbf{U}_{n}^2}\right)^{\mathrm{T}}\left({\mathbf{U}_{n}^1} - {\mathbf{U}_{n}^2}\right){S}_{orth}\mathbf{\Sigma}^2_{\mathbf{G}^{n}_{\mathbf{V}} {\mathbf{G}^{n}_{\mathbf{V}}}^{\mathrm{T}}}\right)\\
		%\leq \delta_{{\mathbf{G}^{n}_{\mathbf{V}} {\mathbf{G}^{n}_{\mathbf{V}}}^{\mathrm{T}}}}^2 \operatorname{tr}\left({S}_{orth}^{\mathrm{T}}\left({\mathbf{U}_{n}^1} - {\mathbf{U}_{n}^2}\right)^{\mathrm{T}}\left({\mathbf{U}_{n}^1} - {\mathbf{U}_{n}^2}\right){S}_{orth}\right) \\
		\leq \left\|\mathbf{G}^{n}_{\mathbf{V}} {\mathbf{G}^{n}_{\mathbf{V}}}^{\mathrm{T}}\right\|_{2}^2 \left\|{\mathbf{U}_{n}^1} - {\mathbf{U}_{n}^2} \right\|_{\mathrm{F}}^{2}
	\end{array} 
\end{equation}
and
\begin{equation}
	\begin{array}{l}
	    \quad \left\|\beta_{n} \mathbf{L}_{n} \left({\mathbf{U}_{n}^1} - {\mathbf{U}_{n}^2}\right) \right\|_{\mathrm{F}}^{2} \\
		=\operatorname{tr}\left(\beta_{n} {\mathbf{L}_{n}}^{\mathrm{T}} \left({\mathbf{U}_{n}^1} - {\mathbf{U}_{n}^2}\right)^{\mathrm{T}}\left({\mathbf{U}_{n}^1} - {\mathbf{U}_{n}^2}\right)\beta_{n}{\mathbf{L}_{n}}\right)  \\
		%= \beta_{n} \operatorname{tr}\left({S}_{orth}^{\mathrm{T}}\left({\mathbf{U}_{n}^1} - {\mathbf{U}_{n}^2}\right)^{\mathrm{T}}\left({\mathbf{U}_{n}^1} - {\mathbf{U}_{n}^2}\right){S}_{orth}\mathbf{\Sigma}^2_{{\mathbf{L}_{n}}}\right) \\
		%\leq \beta_{n} \delta_{{\mathbf{L}_{n}}}^2 \left\|{\mathbf{U}_{n}^1} - {\mathbf{U}_{n}^2} \right\|_{\mathrm{F}}^{2} \\ 
		\leq \beta_{n} \left\|{\mathbf{L}_{n}}\right\|_{2}^2 \left\|{\mathbf{U}_{n}^1} - {\mathbf{U}_{n}^2} \right\|_{\mathrm{F}}^{2}
	\end{array} 
\end{equation}
% the largest eigenvalues
where $\left\|\mathbf{G}_{\mathbf{V}}\right\|_{2}$ and ${\left\|\mathbf{L}_{n}\right\|_{2}}$ are the spectral norm with respect to $\mathbf{G}_{\mathbf{V}}$ and $\mathbf{L}_{n}$. Therefore, ${\nabla_{\mathbf{U}_{n}} \ell(\mathbf{U}_{n})}$ is Lipschitz continuous and the Lipstchitz constant $L_{\mathbf{U}_{n}}$ is bounded. Furthermore, the gradient of temporal regularization satisfies
\begin{equation}
	\begin{array}{l}
	    \quad \left\|\beta_{n} \mathbf{T}_{n}^{\mathrm{T}}\mathbf{T}_{n} \left({\mathbf{U}_{n}^1} - {\mathbf{U}_{n}^2}\right) \right\|_{\mathrm{F}}^{2} 
		%= \beta_{n} \operatorname{tr}\left({S}_{orth}^{\mathrm{T}}\left({\mathbf{U}_{n}^1} - {\mathbf{U}_{n}^2}\right)^{\mathrm{T}}\left({\mathbf{U}_{n}^1} - {\mathbf{U}_{n}^2}\right){S}_{orth}\mathbf{\Sigma}^2_{\mathbf{T}_{n}^{\mathrm{T}}\mathbf{T}_{n}}\right) \\
		\leq \beta_{n} \left\|\mathbf{T}_{n}^{\mathrm{T}}\mathbf{T}_{n}\right\|_{2}^2 \left\|{\mathbf{U}_{n}^1} - {\mathbf{U}_{n}^2} \right\|_{\mathrm{F}}^{2}
	\end{array} 
\end{equation}

Combine with above Equations, we define the Lipschitz constant $L_{\mathbf{U}_{n}}$ as
\begin{equation}
L_{\mathbf{U}_{n}}=\left\{\begin{array}{ll}
\left\|\mathbf{G}^{n}_{\mathbf{V}} {\mathbf{G}^{n}_{\mathbf{V}}}^{\mathrm{T}}\right\|_{2} + \beta_{n} \left\|\mathbf{L}_{n} \right\|_{2}, & \text {Manifold} \\
\left\|\mathbf{G}^{n}_{\mathbf{V}} {\mathbf{G}^{n}_{\mathbf{V}}}^{\mathrm{T}}\right\|_{2} + \beta_{n}  \left\| \mathbf{T}_{n}^{\mathrm{T}}\mathbf{T}_{n} \right\|_{2}, & \text {Temporal} \\
\left\|\mathbf{G}^{n}_{\mathbf{V}} {\mathbf{G}^{n}_{\mathbf{V}}}^{\mathrm{T}}\right\|_{2}, & \text {otherwise }
\end{array}\right. \label{A3}
\end{equation}
This completes the proof. 
\end{proof}

To solve \eqref{Matrix}, we take the derivative and set it to zeros, then we have 
\begin{equation}
\begin{aligned}
   \mathbf{U}_n \longleftarrow \mathcal{P}_{+}\left(\tilde{\mathbf{U}}_n - \frac{1}{L_{\mathbf{U}_n}} \nabla_{\mathbf{U}_n}\ell\left(\tilde{\mathbf{U}}_n\right)\right),
\end{aligned}
\end{equation}
where $\mathcal{P}_{+}(\mathbf{U})$ is the function that projects the negative entries of $\mathbf{U}$ into zeros and $\tilde{\mathbf{U}}_{n}$ is updated by 
\begin{equation*}
\tilde{\mathbf{U}}_{n}^{k} ={\mathbf{U}_{n}^{k}}+\omega_{k}\left({\mathbf{U}_{n}^{k}}-{\mathbf{U}}_{n}^{k-1}\right), \ \text{for} \ k \geq 1.  
\end{equation*}
with the update step size $\omega_{k}$ using \eqref{Step}.

\begin{proof}[\rm{\textbf{Proof of Proposition 2}}]
As in Proposition 1, verifying the convex and Lipschitz continuous properties is straightforward. For the vectorization form, we have
\begin{equation}
\begin{aligned}
   \operatorname{vec}\left(\nabla_{\mathcal{G}} f(\mathcal{G})\right) &= \left(\otimes_{n=N}^{1} \mathbf{U}_{n}^{\mathrm{T}} \mathbf{U}_{n}\right) \operatorname{vec}(\mathcal{G})\\
   &-\left(\otimes_{n=N}^{1} \mathbf{U}_{n}^{\mathrm{T}}\right) \operatorname{vec}(\mathcal{X}),
\end{aligned} \label{GradientCore}
\end{equation}
Then, the Hessian matrix $ \operatorname{vec}\left(\nabla^2_{\mathcal{G}} f(\mathcal{G})\right) = \otimes_{n=N}^{1} \mathbf{U}_{n}^{\mathrm{T}} \mathbf{U}_{n}$, which is positive semidefinite and assures $f(\mathcal{G})$ is convex. Furthermore, we use the properties of Kronecker product to calculate $\nabla_{\mathcal{G}} f(\mathcal{G})$ as follows
\begin{equation}
\begin{aligned}
   \nabla_{\mathcal{G}} f(\mathcal{G}) &= \mathcal{G} \times_{1} {\mathbf{U}_{1}^{\mathrm{T}}\mathbf{U}_{1} \times \cdots \times_{N} \mathbf{U}_{N}^{\mathrm{T}}\mathbf{U}_{N}} \\
   &- \mathcal{X} \times_{1} {\mathbf{U}_{1}^{\mathrm{T}} \times \cdots \times_{N} \mathbf{U}_{N}^{\mathrm{T}}}.
\end{aligned} \label{A6}
\end{equation}
For any given $\mathcal{G}_1$ and $\mathcal{G}_2$, we have 
\begin{equation}
\begin{array}{l}
    \quad \left\|\operatorname{vec}\left(\nabla_{\mathcal{G}} f(\mathcal{G}_1)\right) - \operatorname{vec}\left(\nabla_{\mathcal{G}} f(\mathcal{G}_2)\right)\right\|_{\mathrm{F}} \\
    = \left\| \otimes_{n=N}^{1} \mathbf{U}_{n}^{\mathrm{T}} \mathbf{U}_{n} \left(\operatorname{vec}(\mathcal{G}_1) - \operatorname{vec}(\mathcal{G}_2)\right) \right\|_{\mathrm{F}} \\
	\leq \left\| \otimes_{n=N}^{1} \mathbf{U}_{n}^{\mathrm{T}} \mathbf{U}_{n}\right\|_2 \left\|\operatorname{vec}(\mathcal{G}_1) - \operatorname{vec}(\mathcal{G}_2)\right\|_{\mathrm{F}} \\
	= \prod_{n=1}^{N}\left\|\mathbf{U}_{n}^{\mathrm{T}} \mathbf{U}_{n}\right\|_2 \left\|\operatorname{vec}(\mathcal{G}_1) - \operatorname{vec}(\mathcal{G}_2)\right\|_{\mathrm{F}}.
\end{array} 
\end{equation}
So, the Lipschitz constant of $\nabla_{\mathcal{G}}f(\mathcal{G})$ is $L_\mathcal{G} =\prod_{n=1}^{N}\left\|\mathbf{U}_{n}^{\mathrm{T}} \mathbf{U}_{n}\right\|_2$. This completes the proof. 
\end{proof}	
Based on the results given by Proposition 2, we can use the soft thresholding operator \cite{Xu2015Tucker} to solve the composite model \eqref{Core}, and the result is 
\begin{equation}
    \hat{\mathcal{G}} = T_{L_\mathcal{G}}^{f, g}(\mathcal{G}) = \mathcal{S}_{\frac{\alpha}{L_{\mathcal{G}}}}\left(\tilde{\mathcal{G}}-\frac{1}{L_{\mathcal{G}}} \nabla_{\mathcal{G}} f\left(\tilde{\mathcal{G}}\right)\right), \label{A4}
\end{equation}
where $\mathcal{S}_{\zeta}(\cdot)$ is `shrinkage' operator defining component-wisely as 
\begin{equation*}
    \mathcal{S}_{\mu}(x)=\operatorname{sign}(x) \cdot \max (0,|x|-\mu).
\end{equation*}
and $\tilde{\mathcal{G}}$ is updated by 
\begin{equation*}
\tilde{\mathcal{G}}^{k} ={\mathcal{G}^{k}}+\omega_{k}\left(\mathcal{G}^{k}-\mathcal{G}^{k-1}\right), \ \text{for} \ k \geq 1.  
\end{equation*}
with the update step size \eqref{Step}.
\section{Convergence Analysis}
We provide convergence proof for the proposed algorithm, which is given in the following three steps:

\noindent \textbf{Square summable:} We express \eqref{STRTD} as $\mathbb{F}(\Theta) = \mathbb{F}_1(\Theta) + \mathbb{F}_2(\Theta)$, $\Theta = \{\{\mathbf{U}_n\},\mathcal{G}\}$, where $\mathbb{F}_1$ is either function $\ell$ or $f$ and $\mathbb{F}_2$ is either the $l_1$ norm or a nonnegative projector. The prox-linear updating rule indicates 
\begin{equation}
    \hat{\Theta} = \underset{\Theta}{\operatorname{argmin}} \ \left\langle\nabla_{\Theta} \mathbb{F}_1(\tilde{\Theta}), \Theta -\tilde{\Theta} \right\rangle +\frac{L_{\Theta}}{2}\|\Theta -\tilde{\Theta}\|_{F}^{2} + \mathbb{F}_2(\Theta), \label{A5}
\end{equation}
where $\tilde{\Theta}$ is the extrapolation point. For any $\Theta^k = \{\{\mathbf{U}^k_n\},\mathcal{G}^k\}$ generated by Algorithm \ref{alg1}, it is worth noting that Algorithm \ref{alg1} performs re-update when $\mathbb{F}\left(\Theta_{k}\right) \textless  \ \mathbb{F}\left(\Theta_{k-1}\right)$, which assures the objective $\mathbb{F}$ nonincreasing, i.e., 
\begin{equation}
\begin{aligned}
    \mathbb{F}_1(\Theta^{k}) \leq & \ \mathbb{F}_1(\Theta^{k-1}) + \left\langle\nabla_{\Theta} \mathbb{F}_1(\Theta^{k-1}), \Theta^{k} -\Theta^{k-1}\right\rangle \\ & +\frac{L_{\Theta^{k-1}}}{2}\|\Theta^{k} -\Theta^{k-1}\|_{F}^{2}.
\end{aligned}
\end{equation}
Considering the convexity of $\mathbb{F}_1, \mathbb{F}_2$, we can conclude that
\begin{equation}
\begin{aligned}
    & \mathbb{F}(\Theta^{k}) - \mathbb{F}(\hat{\Theta}) \\
    & \geq \ \frac{L_{\Theta}^{k-1}}{2}\|\hat{\Theta} -{\Theta}^{k-1}\|_{F}^{2} + L_{\Theta}^{k-1} \left\langle {\Theta}^{k-1} - \Theta^{k}, \hat{\Theta} - {\Theta}^{k-1} \right\rangle
\end{aligned}
\end{equation}
Based on the results given by \textit{\textbf{Proposition 1}} and \textit{\textbf{Proposition 2}}, we have $\nabla_{\Theta} \mathbb{F}_1(\Theta)$ is Lipschitz continuous, which has bounded Lipschitz constant. Then for three successive $ \Theta^{k-2}, \Theta^{k-1}, \Theta^{k}$, we have
\begin{equation}
\begin{aligned}
    & \mathbb{F}(\Theta^{k-1}) - \mathbb{F}({\Theta}^k) \\
    & \geq \ \frac{L_{\Theta}^{k-1}}{2}\|\Theta^k -\tilde{\Theta}^{k-1}\|_{F}^{2} + L_{\Theta}^{k-1} \left\langle \tilde{\Theta}^{k-1} - \Theta^{k-1}, \Theta^k  - \tilde{\Theta}^{k-1} \right\rangle \\
    & \geq \ \frac{L_{\Theta}^{k-1}}{2} \|\Theta^{k-1} - \Theta^{k}\|_{F}^{2} - \frac{L_{\Theta}^{k-2} \delta_{\omega}}{2} \|\Theta^{k-2} - \Theta^{k-1}\|_{F}^{2}, \ \delta_{\omega} \textless 1.
\end{aligned}
\end{equation}
Summing the above inequality over $k$ from 1 to $K$, we have 
\begin{equation}
    \mathbb{F}(\Theta^{0}) - \mathbb{F}({\Theta}^K) \geq \sum_{k = 1}^{K}  \text{const.} \ \|\Theta^{k-1} - \Theta^{k}\|_{F}^{2}.
\end{equation}
Letting $K \to \infty $ and observing $\mathbb{F}$ is lower bounded, we have ${\sum_{k=1}^{\infty}\left\|\Theta^{k-1}-\Theta^{k}\right\|_{F}<\infty}$, i.e., 
\begin{equation}
    \lim_{k \to \infty}\left(\Theta^{k}-\Theta^{k-1}\right) = 0
\end{equation}
\noindent \textbf{Subsequence convergence:} Recall the prox-linear operator mentioned in \eqref{A5}, which is a convex minimization problem. Depending on the square summable property, we set $\hat{\Theta}$ as a limit point of $ \Theta $. Recall that
\begin{equation}
    \hat{\Theta} = \underset{\Theta}{\operatorname{argmin}} \ \left\langle\nabla_{\Theta} \mathbb{F}_1(\hat{\Theta}), \Theta -\hat{\Theta} \right\rangle +\frac{\hat{L}_{\Theta}}{2}\|\Theta -\hat{\Theta}\|_{F}^{2} + \mathbb{F}_2(\Theta)
\end{equation}
Hence, $\hat{\Theta}$ satisfies the first-order optimality condition of \eqref{STRTD}
\begin{equation}
    \left\langle\nabla_{\Theta} \ell(\Theta) + \mathbb{\lambda} \mathbb{P},  \Theta - \hat{\Theta} \right\rangle \geq \mathbf{0}, \ \text{for \ all } \Theta_k, \ \text{some} \ \mathbb{P} \in \partial \mathbb{F}_2.
\end{equation}
Then there is a subsequence $\Theta^k = \{\{\mathbf{U}^k_n\},\mathcal{G}^k\}$ converging to $\hat{\Theta}$ and $\hat{\Theta}$ is a stationary point.

\noindent \textbf{Global convergence:} Guided by \cite{PALM2014}, it is straightforward to demonstrate that $\mathbb{F}$ satisfies the Kurdyka–Lojasiewicz (KL) property at $\hat{\Theta}$, namely, there exist ${\mu, \rho>0, \eta \in [0,1]}$, and a neighborhood ${\mathcal{B}(\hat{\Theta}, \rho) = \left\{\Theta:\|\Theta - \hat{\Theta}\|_{F}^2 \leq \rho\right\}}$ such that
\begin{equation}
|\mathbb{F}(\Theta)-\mathbb{F}(\hat{\Theta})|^{\eta} \leq \mu \cdot \operatorname{dist}(\mathbf{0}, \partial \mathbb{F}(\Theta)), \text { for \ all } \Theta \in \mathcal{B}(\hat{\Theta}, \rho).
\end{equation}
Combining the subsequence convergence and KL property, the sequence $\Theta^k$ converges to $\hat{\Theta}$, which is a critical point of equation \eqref{STRTD}.

\section{Computational Complexity Analysis}
The Tucker decomposition algorithms compute the huge matrix multiplication and suffer from very high computational complexity; we combine the low-rank approximation with population Tucker decomposition strategies to reduce the computational complexity \cite{Zhou2015}. Here, we analyze the computational complexity of the proposed STRTD. Suppose that ${\mathcal{X} \in \mathbb{R}^{I_{1} \times \ldots \times I_{N}}}$ and the core tensor ${\mathcal{G} \in \mathbb{R}^{I_{1} \times \ldots \times I_{N}}}$, we have the basic computational complexity: the computational cost of $ {\mathbf{U}_{n}^{\mathrm{T}}\mathbf{U}_{n}} $ is $\mathcal{O}(I_{n}^3)$ and the mode-n product with the matrix ${\mathbf{U}_{n}}$ of tensor ${\mathcal{G}}$ is ${\mathcal{O}(\sum_{n=1}^N \prod_{i=1}^{n} I_{i} \prod_{j=1}^{N} I_{j} )}$. Furthermore, we reformulate the Kronecker product in ${\mathbf{G}^{n}_{\mathbf{V}}} = {\mathbf{G}_{(n)}} \mathbf{V}_{n}^{\mathrm{T}}$ but let 
\begin{equation*}
\mathcal{Y} ={\mathcal{G}} \times_{1}{\mathbf{U}_{1}} \cdots \times_{n-1}{\mathbf{U}_{n-1}}\times_{n+1}{\mathbf{U}_{n+1}} \cdots \times_{N}{\mathbf{U}_{N}},
\end{equation*}
such that we have ${\mathbf{G}^{n}_{\mathbf{V}}} = \mathcal{Y}_{(n)}$ and its computational cost is
\begin{equation}
\begin{aligned}
\mathcal{O}\left(\mathbf{G}^{n}_{\mathbf{V}}\right)\ &= \ \mathcal{O}\left(\sum_{j=1}^{n-1}\left(\prod_{i=1}^{j} I_{i}\right)\left(\prod_{i=j}^{N} I_{i}\right)\right) + \\
&\quad \mathcal{O}\left(\left(\prod_{i=1}^{n} I_{i}\right) \sum_{j=n+1}^{N}\left(\prod_{i=n+1}^{j} I_{i}\right)\left(\prod_{i=j}^{N} I_{i}\right)\right)\\
&\leq \mathcal{O}\left(\sum_{n=1}^{N}\left(\prod_{i=1}^{n} I_{i}\right)\left(\prod_{j=n}^{N} I_{j}\right)\right)
\end{aligned} \label{A7}
\end{equation}
Also, we conclude that the computational cost of tensor unfolding, soft-thresholding operator, and projection to nonnegative is negligible compared to gradient computing. 

Considering the proposed APG-based optimization for core tensor ``shrinkage'', the computation of $\nabla_\mathcal{G} f\left(\mathcal{G}\right)$ requires
\begin{equation}
\mathcal{O}\left(\sum_{n=1}^{N} I_{n}^{3} +\sum_{n=1}^{N} I_{n} \prod_{i=1}^{N} I_{i}+\sum_{n=1}^{N}\left(\prod_{i=1}^{n} I_{i}\right)\left(\prod_{j=n}^{N} I_{j}\right)\right). \label{A8}
\end{equation}
where the first part comes from the computation of all $\mathbf{U}_{n}^{\mathrm{T}}\mathbf{U}_{n}$, and the second and third parts are respectively, from the computations of the first and second
terms in \eqref{A6}. 

Similarly, we use \eqref{A7} to calculate the computational complexity of $\nabla_{\mathbf{U}_{n}} \ell(\mathbf{U}_{n})$ and requires
\begin{equation}
\mathcal{O}\left(I_{n}\left(\prod_{i=1}^{n} I_{i}\right) + I_{n}^3\right) + \mathcal{O}\left(\prod_{i=1}^{n} I_{i}\right) + \mathcal{O}\left(I^3_{n}\right) + \mathcal{O}\left(\mathbf{G}^{n}_{\mathbf{V}}\right). \label{A9}
\end{equation}
The first three parts are from the computations of the three terms in \eqref{A1}, and \eqref{A9} is dominated by the last part. So, the computational cost of $\nabla_\mathcal{G} f\left(\mathcal{G}\right)$ and $\nabla_{\mathbf{U}_{n}} \ell(\mathbf{U}_{n})$ are
\begin{equation}
\mathcal{O}\left(\sum_{n=1}^{N}\left(\prod_{i=1}^{n} I_{i}\right)\left(\prod_{j=n}^{N} I_{j}\right)\right). 
\end{equation}
}

\bibliographystyle{IEEEtran}
\bibliography{ref.bib}

\begin{thebibliography}{10}
\providecommand{\url}[1]{#1}
\csname url@samestyle\endcsname
\providecommand{\newblock}{\relax}
\providecommand{\bibinfo}[2]{#2}
\providecommand{\BIBentrySTDinterwordspacing}{\spaceskip=0pt\relax}
\providecommand{\BIBentryALTinterwordstretchfactor}{4}
\providecommand{\BIBentryALTinterwordspacing}{\spaceskip=\fontdimen2\font plus
\BIBentryALTinterwordstretchfactor\fontdimen3\font minus
  \fontdimen4\font\relax}
\providecommand{\BIBforeignlanguage}[2]{{%
\expandafter\ifx\csname l@#1\endcsname\relax
\typeout{** WARNING: IEEEtran.bst: No hyphenation pattern has been}%
\typeout{** loaded for the language `#1'. Using the pattern for}%
\typeout{** the default language instead.}%
\else
\language=\csname l@#1\endcsname
\fi
#2}}
\providecommand{\BIBdecl}{\relax}
\BIBdecl

\bibitem{CHEN2018patterns}
X.~Chen, Z.~He, and J.~Wang, ``Spatial-temporal traffic speed patterns
  discovery and incomplete data recovery via {SVD}-combined tensor
  decomposition,'' \emph{Transportation Research Part C: Emerging
  Technologies}, vol.~86, pp. 59--77, 2018.

\bibitem{imputeTS}
S.~Moritz and T.~Bartz-Beielstein, ``{imputeTS: Time Series Missing Value
  Imputation in {R}},'' \emph{{The R Journal}}, vol.~9, no.~1, pp. 207--218,
  2017.

\bibitem{Thomas2021}
T.~Thomas and E.~Rajabi, ``A systematic review of machine learning-based
  missing value imputation technique,'' \emph{Data Technologies and
  Applicationse}, 2021.

\bibitem{TAN201315}
H.~Tan, G.~Feng, J.~Feng, W.~Wang, Y.-J. Zhang, and F.~Li, ``A tensor-based
  method for missing traffic data completion,'' \emph{Transportation Research
  Part C: Emerging Technologies}, vol.~28, pp. 15--27, 2013.

\bibitem{Chen20219548664}
X.~Chen, M.~Lei, N.~Saunier, and L.~Sun, ``Low-rank autoregressive tensor
  completion for spatiotemporal traffic data imputation,'' \emph{IEEE
  Transactions on Intelligent Transportation Systems}, pp. 1--10, 2021.

\bibitem{Kolda2009}
T.~G. Kolda and B.~W. Bader, ``Tensor decompositions and application,''
  \emph{SIAM Review}, vol.~5, no.~3, p. 455–500, 2009.

\bibitem{Song2019}
Q.~Song, H.~Ge, J.~Caverlee, and X.~Hu, ``Tensor completion algorithms in big
  data analytics,'' \emph{ACM Transactions on Knowledge Discovery from Data},
  vol.~13, no.~1, p.~48, 2019.

\bibitem{CHEN2019patterns}
X.~Chen, Z.~He, Y.~Chen, Y.~Lu, and J.~Wang, ``Missing traffic data imputation
  and pattern discovery with a bayesian augmented tensor factorization model,''
  \emph{Transportation Research Part C: Emerging Technologies}, vol. 104, pp.
  66--77, 2019.

\bibitem{Rose2014}
M.~T. Bahadori, Q.~R. Yu, and Y.~Liu, ``Fast multivariate spatio-temporal
  analysis via low rank tensor learning,'' in \emph{Neural Information
  Processing Systems (NIPS)}, 2014, p. 3491–3499.

\bibitem{SaidSpatiotemporal}
A.~B. Said and A.~Erradi, ``Spatiotemporal tensor completion for improved urban
  traffic imputation,'' \emph{IEEE Transactions on Intelligent Transportation
  Systems}, pp. 1--14, 2021.

\bibitem{Roughan2012}
M.~Roughan, Y.~Zhang, W.~Willinger, and L.~Qiu, ``Spatio-temporal compressive
  sensing and internet traffic matrices (extended version),'' \emph{IEEE/ACM
  Transactions on Networking}, vol.~20, no.~3, pp. 662--676, 2012.

\bibitem{TASTWang}
Y.~Wang, Y.~Zhang, X.~Piao, H.~Liu, and K.~Zhang, ``Traffic data reconstruction
  via adaptive spatial-temporal correlations,'' \emph{IEEE Transactions on
  Intelligent Transportation Systems}, vol.~20, no.~4, pp. 1531--1543, 2019.

\bibitem{LRHTWang2021}
X.~Wang, Y.~Wu, D.~Zhuang, and L.~Sun, ``Low-rank {H}ankel tensor completion
  for traffic speed estimation,'' \emph{IEEE Transactions on Intelligent
  Transportation Systems}, vol.~24, no.~5, pp. 4862--4871, 2023.

\bibitem{Xutao7460200}
X.~Li, M.~K. Ng, G.~Cong, Y.~Ye, and Q.~Wu, ``{MR-NTD}: Manifold regularization
  nonnegative {T}ucker decomposition for tensor data dimension reduction and
  representation,'' \emph{IEEE Transactions on Neural Networks and Learning
  Systems}, vol.~28, no.~8, pp. 1787--1800, 2017.

\bibitem{WuTrac2020}
P.~Wu, L.~Xu, and Z.~Huang, ``Imputation methods used in missing traffic data:
  A literature review,'' in \emph{Artificial Intelligence Algorithms and
  Applications}, 2020, pp. 662--677.

\bibitem{yu2016temporalr}
H.-F. Yu, N.~Rao, and I.~S. Dhillon, ``Temporal regularized matrix
  factorization for high-dimensional time series prediction,'' in \emph{Neural
  Information Processing Systems (NIPS)}, 09 2016.

\bibitem{Candes2009}
E.~J. Candes and B.~Recht, ``Exact matrix completion via convex optimization,''
  \emph{Foundations of Computational Mathematics}, vol.~9, no.~6, p. 717–772,
  2009.

\bibitem{Liu6138863}
J.~Liu, P.~Musialski, P.~Wonka, and J.~Ye, ``Tensor completion for estimating
  missing values in visual data,'' \emph{IEEE Transactions on Pattern Analysis
  and Machine Intelligence}, vol.~35, no.~1, pp. 208--220, 2013.

\bibitem{RAN201654}
B.~Ran, H.~Tan, Y.~Wu, and P.~J. Jin, ``Tensor based missing traffic data
  completion with spatial–temporal correlation,'' \emph{Physica A :
  Statistical Mechanics and its Applications}, vol. 446, pp. 54--63, 2016.

\bibitem{Huachun2014}
H.~Tan, J.~Feng, Z.~Chen, F.~Yang, and W.~Wang, ``Low multilinear rank
  approximation of tensors and application in missing traffic data,''
  \emph{Advances in Mechanical Engineering}, vol.~6, pp. 1575--1597, 2014.

\bibitem{Yokota8578959}
T.~Yokota, B.~Erem, S.~Guler, S.~K. Warfield, and H.~Hontani, ``Missing slice
  recovery for tensors using a low-rank model in embedded space,'' in
  \emph{IEEE/CVF Conference on Computer Vision and Pattern Recognition (CVPR)},
  2018, pp. 8251--8259.

\bibitem{PanLRSETD}
C.~Pan, C.~Ling, H.~He, L.~Qi, and Y.~Xu, ``Low-rank and sparse enhanced
  {T}ucker decomposition for tensor completion,'' \emph{arXiv}, vol.
  abs/2010.00359, 2020.

\bibitem{BGCPChen2019}
X.~Chen, Z.~He, and L.~Sun, ``A bayesian tensor decomposition approach for
  spatiotemporal traffic data imputation,'' \emph{Transportation Research Part
  C: Emerging Technologies}, vol.~98, pp. 73--84, 2019.

\bibitem{CHEN2021Tubal}
X.~Chen, Y.~Chen, N.~Saunier, and L.~Sun, ``Scalable low-rank tensor learning
  for spatiotemporal traffic data imputation,'' \emph{Transportation Research
  Part C: Emerging Technologies}, vol. 129, p. 103226, 2021.

\bibitem{shi2020block}
Q.~Shi, J.~Yin, J.~Cai, A.~Cichocki, T.~Yokota, L.~Chen, M.~Yuan, and J.~Zeng,
  ``Block {H}ankel tensor {ARIMA} for multiple short time series forecasting,''
  in \emph{AAAI Conference on Artificial Intelligence (AAAI)}, vol.~30, no.~04,
  2020, pp. 5758--5766.

\bibitem{stTT2021}
Z.~Zhang, Y.~Chen, H.~He, and L.~Qi, ``A tensor train approach for internet
  traffic data completion,'' \emph{Annals of Operations Research}, vol.~06, pp.
  12--19, 2021.

\bibitem{Wu8421043}
Y.~Wu, H.~Tan, Y.~Li, J.~Zhang, and X.~Chen, ``A fused {CP} factorization
  method for incomplete tensors,'' \emph{IEEE Transactions on Neural Networks
  and Learning Systems}, vol.~30, no.~3, pp. 751--764, 2019.

\bibitem{Goulart2017}
J.~H. de~Morais~Goulart and G.~Favier, ``Low-rank tensor recovery using
  sequentially optimal modal projections in iterative hard thresholding,''
  \emph{SIAM Journal on Scientific Computing}, vol.~39, no.~3, pp. 860--889,
  2017.

\bibitem{ZHANG2019337}
H.~Zhang, P.~Chen, J.~Zheng, J.~Zhu, G.~Yu, Y.~Wang, and H.~X. Liu, ``Missing
  data detection and imputation for urban {ANPR} system using an iterative
  tensor decomposition approach,'' \emph{Transportation Research Part C:
  Emerging Technologies}, vol. 107, pp. 337--355, 2019.

\bibitem{Wang8708929}
J.~Wang, J.~Wu, Z.~Wang, F.~Gao, and Z.~Xiong, ``Understanding urban dynamics
  via context-aware tensor factorization with neighboring regularization,''
  \emph{IEEE Transactions on Knowledge and Data Engineering}, vol.~32, no.~11,
  pp. 2269--2283, 2020.

\bibitem{Goulart201701}
J.~H. Goulart, A.~Kibangou, and G.~Favier, ``Traffic data imputation via tensor
  completion based on soft thresholding of {T}ucker core,''
  \emph{Transportation Research Part C Emerging Technologies}, vol.~85, pp.
  348--362, 12 2017.

\bibitem{Chen2022patterns}
X.~Chen, C.~Zhang, X.~Chen, N.~Saunier, and L.~Sun, ``Discovering dynamic
  patterns from spatiotemporal data with time-varying low-rank
  autoregression,'' \emph{arXiv}, vol. abs/2211.15482, 2022.

\bibitem{GongARTD}
W.~Gong, Z.~Huang, and L.~Yang, ``Accurate regularized tucker decomposition for
  image restoration,'' \emph{Applied Mathematical Modeling}, vol. 123, no.~11,
  pp. 75--86, 2023.

\bibitem{ChenSTDC}
Y.-L. Chen, C.-T. Hsu, and H.-Y.~M. Liao, ``Simultaneous tensor decomposition
  and completion using factor priors,'' \emph{IEEE Transactions on Pattern
  Analysis and Machine Intelligence}, vol.~36, no.~3, pp. 577--591, 2014.

\bibitem{YuSBCD}
Q.~Yu, X.~Zhang, Y.~Chen, and L.~Qi, ``Low tucker rank tensor completion using
  a symmetric block coordinate descent method,'' \emph{Numerical Linear Algebra
  with Applications}, vol.~30, no.~3, p. e2464, 2023.

\bibitem{Auxiliary2012}
A.~Narita, K.~Hayashi, R.~Tomioka, and H.~Kashima, ``Tensor factorization using
  auxiliary information,'' \emph{Data Mining and Knowledge Discovery}, vol.~25,
  p. 298–324, 2012.

\bibitem{Sinha_2022_WACV}
T.~K. Sinha, J.~Naram, and P.~Kumar, ``Nonnegative low-rank tensor completion
  via dual formulation with applications to image and video completion,'' in
  \emph{IEEE/CVF Winter Conference on Applications of Computer Vision (WACV)},
  January 2022, pp. 3732--3740.

\bibitem{Xu2012L2}
Y.~Xu and W.~Yin, ``A {B}lock {C}oordinate {D}escent method for regularized
  multiconvex optimization with applications to nonnegative tensor
  factorization and completion,'' \emph{SIAM Journal on Imaging Sciences},
  vol.~6, no.~3, pp. 1758--1789, 2013.

\bibitem{Xu2015Tucker}
Y.~Xu, ``Alternating proximal gradient method for sparse nonnegative {T}ucker
  decomposition,'' \emph{Mathematical Programming Computation}, vol.~5, no.~3,
  p. 455–500, 2015.

\bibitem{FISTA2022}
J.~Liang, T.~Luo, and C.-B. Sch\"{o}nlieb, ``Improving “fast iterative
  shrinkage-thresholding algorithm”: {F}aster, smarter, and greedier,''
  \emph{SIAM Journal on Scientific Computing}, vol.~44, no.~3, pp.
  A1069--A1091, 2022.

\bibitem{tSVD2017}
Z.~Zhang and S.~Aeron, ``Exact tensor completion using t-svd,'' \emph{IEEE
  Transactions on Signal Processing}, vol.~65, no.~6, pp. 1511--1526, 2017.

\bibitem{XieKBR2}
Q.~Xie, Q.~Zhao, D.~Meng, and Z.~Xu, ``Kronecker-basis-representation based
  tensor sparsity and its applications to tensor recovery,'' \emph{IEEE
  Transactions on Pattern Analysis and Machine Intelligence}, vol.~40, no.~8,
  pp. 1888--1902, 2018.

\bibitem{Tatsuya2022}
R.~Yamamoto, H.~Hontani, A.~Imakura, and T.~Yokota, ``Fast algorithm for
  low-rank tensor completion in delay-embedded space,'' in \emph{2022 IEEE/CVF
  Conference on Computer Vision and Pattern Recognition (CVPR)}, 2022, pp.
  2048--2056.

\bibitem{Chen2022Prediction}
X.~Chen and L.~Sun, ``Bayesian temporal factorization for multidimensional time
  series prediction,'' \emph{IEEE Transactions on Pattern Analysis and Machine
  Intelligence}, vol.~44, no.~9, pp. 4659--4673, 2022.

\bibitem{PALM2014}
J.~Bolte, S.~Sabach, and M.~Teboulle, ``Proximal alternating linearized
  minimization for nonconvex and nonsmooth problems,'' \emph{Mathematical
  Programming}, vol. 146, no.~7, p. 459–494, 2014.

\bibitem{Zhou2015}
G.~Zhou, A.~Cichocki, Q.~Zhao, and S.~Xie, ``Efficient nonnegative {T}ucker
  decompositions: Algorithms and uniqueness,'' \emph{IEEE Transactions on Image
  Processing}, vol.~24, no.~12, pp. 4990--5003, 2015.

\end{thebibliography}

\end{document}